\documentclass[11pt]{article}
\usepackage[preprint]{acl}

\usepackage{times}
\usepackage{latexsym}
\usepackage{makecell}
\usepackage{amsmath}
\usepackage{amssymb}
\usepackage{colortbl}   
\usepackage{enumitem}
\definecolor{lightgreen}{RGB}{220, 240, 220} 
\usepackage{float}
\usepackage{booktabs}

\usepackage[T1]{fontenc}

\usepackage[utf8]{inputenc}

\usepackage{microtype}

\usepackage{inconsolata}

\usepackage{graphicx}

\usepackage{multirow}
\usepackage{colortbl}
\usepackage{adjustbox}
\usepackage{xcolor}

\author{
    \textbf{Jianuo Huang$^{1,2}$\thanks{Equal contribution. $^\dag$Corresponding author.} \quad Yaojie Zhang$^{1,3*}$ \quad Yicun Yang$^{1}$ \quad Benhao Huang$^{4}$} \vspace{1pt}\\
    \textbf{Biqing Qi$^{5}$ \quad Dongrui Liu$^{5}$ \quad Linfeng Zhang$^{1\dag}$}\vspace{3pt} \\
    $^1$School of Artificial Intelligence, Shanghai Jiao Tong University \\ \vspace{1pt}
    $^2$Huazhong University of Science and Technology \\ \vspace{1pt}
    $^3$University of Electronic Science and Technology of China \\ \vspace{1pt}
    $^4$Carnegie Mellon University
    $^5$Shanghai Artificial Intelligence Laboratory
     \vspace{-3pt} \\
    \texttt{\small \{jianuohuang82,yaojiezhang288\}@gmail.com} \vspace{5pt} \\
}

\title{Mask Tokens as Prophet: Fine-Grained Cache Eviction \\for Efficient dLLM Inference}

\newcommand{\mymethod}{MaskKV}

\begin{document}
\maketitle

\begin{abstract}
Diffusion large language models (dLLMs) present a promising alternative to dominant autoregressive models (ARMs) by the ability of parallel decoding at the expense of substantial computation and memory costs. 
Specifically, the cache mechanism for bidirectional attention in dLLMs demands large memory footprint, restricting their ability to handle long contexts under resource-limited settings. 
Existing cache eviction strategies are designed for ARMs and ignore the unique characteristics of dLLMs, thus leading to unsatisfactory performance.
To address these challenges, we introduce \emph{MaskKV}, a training-free cache eviction framework tailored to dLLMs,  focusing on the effect of mask tokens in dLLMs. 
\emph{MaskKV} is built on two key innovations: 
(1) a mask-query guided scoring mechanism that leverages attention weights to identify and evict less critical prompt tokens for each head; 
(2) an adaptive cache budgeting strategy that improves efficiency by reducing allocation in intermediate layers and concentrating resources on prompt-preferring heads. 
On LLaDA with \emph{MaskKV}, compressing the KV cache to only 256 pairs (less than 5\% of tokens) retains 94\% of the full-cache performance on LongBench and achieves up to 31 $\times$ acceleration at 32k prompt length. 
\emph{The code is publicly available as an open-source project.}\footnote{\url{https://github.com/jianuo-huang/MaskKV}}
\end{abstract}
\section{Introduction}

\begin{figure*}[t]
  \centering
  \includegraphics[width=1.0\linewidth]{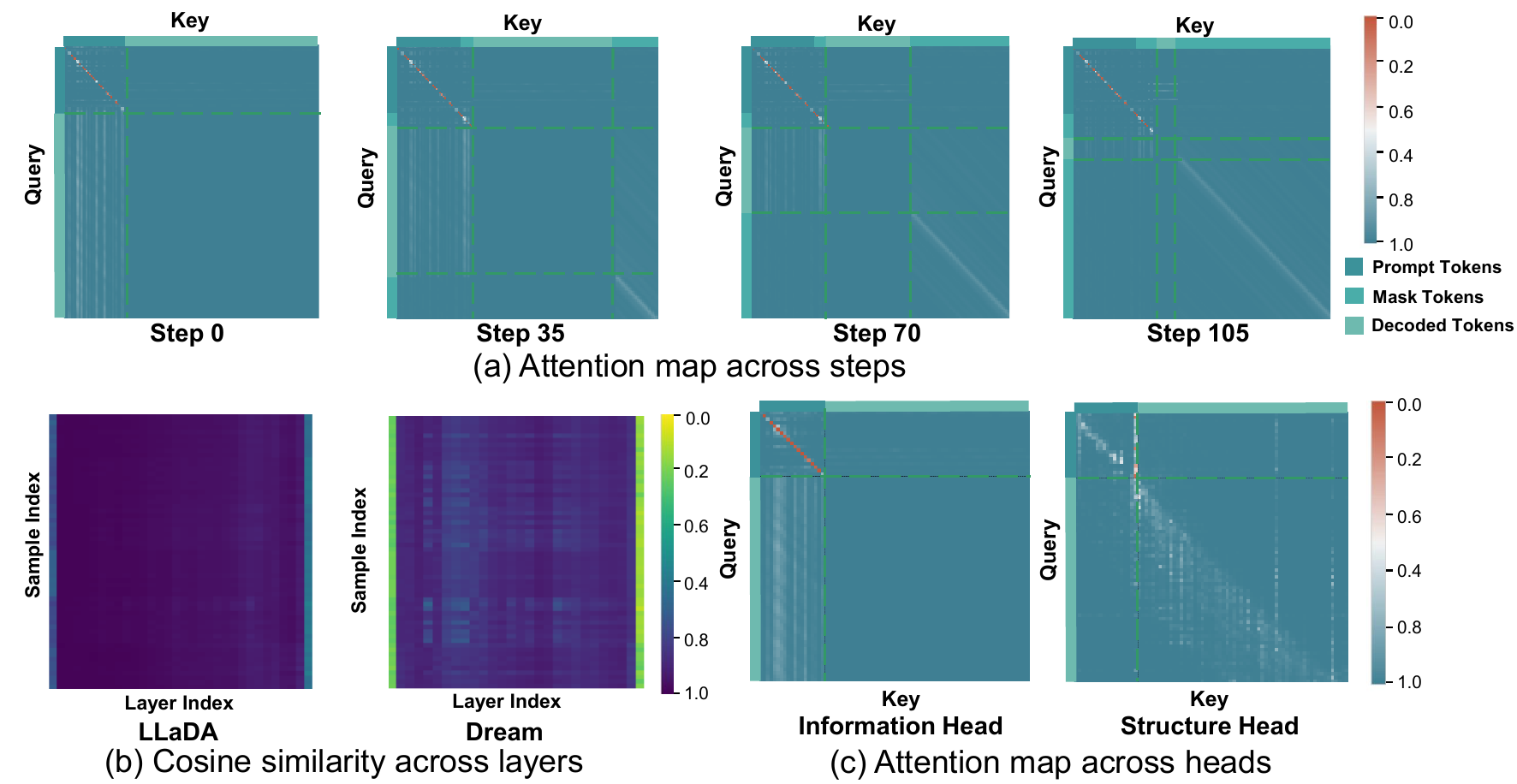}
  \caption{
  \textbf{Visualization on attention maps and features in LLaDA.}
    (a) The queries from the mask token tend to clearly concentrate on several ``important'' prompt tokens, indicating they are able to choose the valuable prefix. (b) The first layer and the last layer in diffusion LLMs tend to show significantly lower cosine similarity compared to their adjacent layers, indicating these two layers contribute more than other layers in generation. (c) ``information heads'' tend to focus more on the previous prompts while ``structure heads'' focus more on the mask tokens.
    }
  \label{fig:attention_map}
\end{figure*}

Over the past few years, autoregressive language models have dominated text generation~\citep{zhao2023survey}, but their strictly left-to-right decoding enforces sequential inference and limits throughput. Diffusion large language models (dLLMs) lift the latency ceiling by iteratively denoising a fully masked sequence, enabling parallel prediction of all tokens with bidirectional attention for richer contextual reasoning~\citep{genimid,chen2025dpad, wang2025diffusion}. Closed-source systems such as Gemini Diffusion~\citep{genimid} and Mercury~\citep{khanna2025mercury} have already pushed this paradigm to production scale, sustaining thousand-token-per-second decoding and proving its commercial viability. Open-source counterparts, LLaDA-8B (trained from scratch)~\citep{nie2025large} and Dream-7B (initialized from AR)~\citep{ye2025dream}, perform comparably to ARMs of similar scale on downstream tasks, confirming diffusion as a competitive architecture for text generation.

To reduce redundant computation in the iterative denoising process, cache mechanisms tailored for bidirectional attention have been proposed~\citep{liu2025dllm, wu2025fast, ma2025dkv, hu2025accelerating}. 
Unlike ARMs, which only maintain and reuse key--value (KV) states for past tokens, dLLMs must recompute and cache features for the entire sequence at every denoising step, including the input prompt, generated tokens, and masked tokens. 
This design amplifies both the storage and update cost of caches, introducing significant memory and runtime overhead~\citep{hu2025accelerating}. 
While eviction strategies have been extensively studied in ARMs~\citep{zhang2023h2o,li2024snapkv}, they largely depend on causal attention over past tokens and thus cannot be directly applied to dLLMs, where bidirectional attention also accesses undecoded positions. 
The few existing efforts for dLLMs mainly focus on block diffusion~\citep{wu2025fast,song2025sparse}, but these approaches sacrifice parallelism and require repeated attention computations, undermining potential acceleration.

These differences between ARMs and dLLMs render existing eviction strategies ineffective and necessitate a rethinking of cache eviction tailored to pure diffusion architectures. This calls for a systematic study of dLLMs to revisit two essential problems: identifying which tokens are most critical to preserve and determining how to allocate cache budgets across layers and heads.

\noindent \textbf{Which tokens should be evicted?}
 Due to the causal attention and the paradigm of ``next-token-prediction, ARMs can not directly formulate the next multiple tokens. As a result, the cache eviction methods in ARMs usually rely only on past tokens (\emph{i.e.}, the prompt tokens and generated tokens).
In contrast, dLLMs can access undecoded positions (\emph{i.e.,} the mask tokens), which may bring new possibilities.
As shown in Fig.~\ref{fig:attention_map} (a), we find that the past tokens in dLLMs exhibit strong locality, primarily attending to themselves and nearby neighbors due to positional bias. 
In contrast, masked tokens maintain stable attention across denoising steps and consistently highlight a small set of pivotal prompt tokens, making the attention scores from the mask tokens a good metric for identifying the crucial tokens in the prompts during the entire decoding process.
Based on this observation, we propose \emph{Mask-Voting} to leverage the attention scores from mask tokens to identify pivotal prompt tokens and safely evict less important cache.

\noindent \textbf{How to allocate the cache budget?} 
Existing ARM cache-budget schemes distribute per-layer and per-head capacity based on attention over past tokens  ~\citep{wang2024squeezeattention,feng2024ada}, but the presence of masked tokens in dLLMs disrupts these patterns, demanding a new allocation strategy.
As shown in Fig.~\ref{fig:attention_map} (b), dLLMs show  clear layer-wise differences in importance: Features exhibit significant change after passing the first and last layers, while having very minimal change in the middle layers, indicating the first and last layers are indispensable while the middle layers tend to be redundant. 
At the head level, as shown in Fig.~\ref{fig:attention_map} (c),  bidirectional attention yields specialized heads, as some perform prompt‑based information extraction while others focus on masked‑token structural planning.
To capture head-level reliance on prompts, we introduce a \emph{prompt-preference} metric based on mask-query attention.
Considering both importance and informativeness, we propose a two-stage budget allocation scheme that first assigns the KV cache budget across layers by importance and then refines the allocation across heads using the prompt-preference metric, avoiding allocation of KV to mask-dominated layers and heads.
Based on the above observations, this paper introduces \emph{MaskKV} as a KV eviction framework tailored to dLLMs, with the following contributions: 
\begin{enumerate}
\item
We analyze the attention behaviors of dLLMs and revealed several useful insights for KV cache eviction, showing how the mask tokens can participate in the judgment of important KV and the allocation of KV budgets.

\item 
We introduce  \emph{MaskKV}, a KV cache eviction framework tailored to diffusion LLMs, which is composed of the mask-query guided token eviction, offline layer-wise budget allocation, and adaptive head-wise budget redistribution.
\item Extensive experiments on LongBench with LLaDA and Dream show that \emph{MaskKV} substantially reduces memory and computation overhead while preserving accuracy. Specifically, on LLaDA, it reaches 94\% of the full-cache performance with the KV cache compressed to only 256 pairs.
\end{enumerate}

\section{Related Work}
\subsection{Diffusion Models for Language}
Diffusion large language models (dLLMs) have emerged as a compelling non-autoregressive paradigm for text generation~\citep{li2025survey}. Their core mechanism involves a progressive refinement process, where a sequence is generated by iteratively denoising a noise-corrupted input over a series of discrete steps.
Recent works have demonstrated both the scalability of this architecture ~\citep{nie2025large} and the effectiveness of training techniques such as AR-based initialization and context-adaptive noise scheduling~\citep{ye2025dream}, achieving competitive performance.

The development of diffusion LLMs is now driven by the twin objectives of accelerating inference and improving generation quality.
To accelerate inference, Fast-dLLM~\citep{wu2025fast} introduces a block-wise approximate KV cache tailored for dLLM, while dLLM-Cache ~\citep{liu2025dllm} leverages feature caching to reduce redundant computation. SlowFast Sampling~\citep{wei2025accelerating} jointly considers confidence, convergence and position  for dynamic decoding, achieving a practical trade-off between speed and quality.
For enhancing performance, DEADAL~\citep{li2025beyond} addressed the limitation of fixed-length generation by introducing variable-length denoising.
LongLLaDA~\citep{liu2025longllada} extends diffusion LLMs to long contexts. These advancements have driven the demand for longer-context handling, which under caching mechanisms leads to prohibitive memory overhead.

\subsection{KV Cache Compression}
The substantial memory footprint of the Key-Value (KV) cache presents a primary bottleneck for long-context inference in Large Language Models (LLMs). 
Early approaches such as LongFormer~\citep{beltagy2020longformer} and StreamingLLM~\citep{xiao2023efficient} used static, content-agnostic rules (e.g., sliding windows, attention sinks), but their tendency to drop long-range information spurred interest in content-aware, dynamic eviction policies~\citep{zhang2023h2o, li2024snapkv, devoto2024simple}. 
Prominent examples include H2O~\citep{zhang2023h2o}, which greedily evicts KVs with low historical attention scores; SnapKV~\citep{li2024snapkv}, which filters for key KVs during the prefill phase.
Building on the recognition of specialized model components, recent studies have explored finer-grained and adaptive budget allocation strategies~\citep{cai2024pyramidkv,wang2024squeezeattention,feng2024ada,xiao2024duoattention}. 
At the layer level, methods either vary cache sizes across layers or dynamically reallocate budgets according to prompt-specific importance~\citep{cai2024pyramidkv,wang2024squeezeattention}. 
At the head level, approaches assign differentiated budgets or categorize heads into functional roles for selective caching~\citep{feng2024ada,xiao2024duoattention}. 
While effective for ARMs, these techniques depend on sequential decoding and are incompatible with the iterative inference of dLLMs. 
Existing attempts, such as Sparse-dLLM, provide only preliminary insights into sparse caching in dLLMs. 
We thus propose a more comprehensive and fine-grained framework for KV cache eviction for dLLMs.

\section{Methodology}

\begin{figure*}[t]
  \centering
  \includegraphics[width=1.0\linewidth]{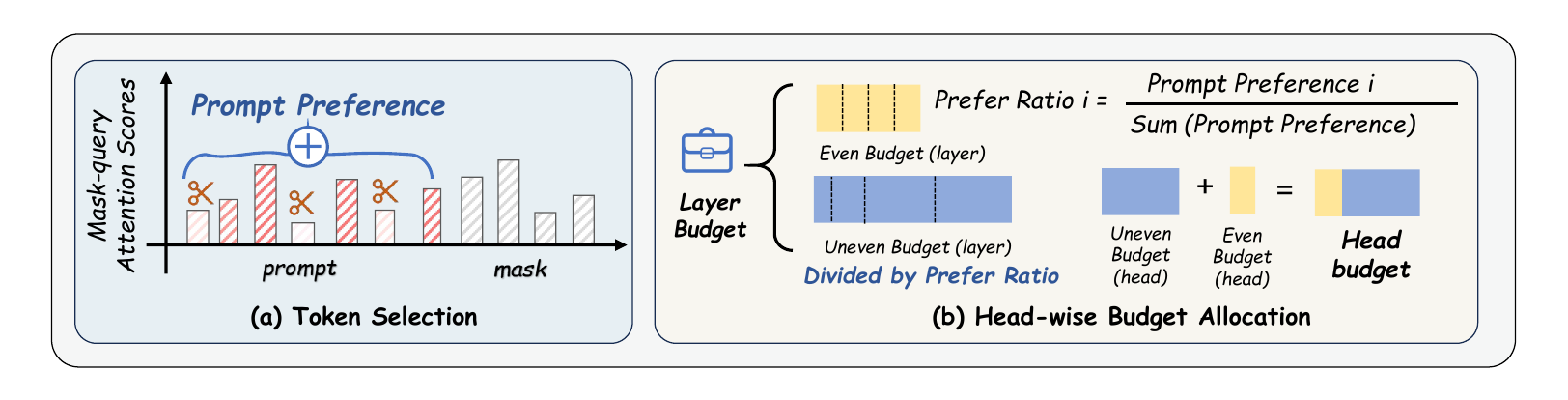}
  \caption{\textbf{The \mymethod{} pipeline.} We use Mask Voting to assess token importance, then apply adaptive budget allocation where layers and heads receive budgets by boundary awareness and prompt preference. Tokens are evicted based on their importance under the given budget.}
  \label{fig:pipline}
\end{figure*}

\subsection{Preliminaries}

\noindent \textbf{Inference of Diffusion Large Language Models.}
Diffusion language models (dLLMs) generate text through an iterative unmasking process over T discrete decoding steps, progressively refining a fully masked sequence into the final output.
We first define the token vocabulary as $\mathcal{T}$, which includes a special mask token $\texttt{[MASK]} \in \mathcal{T}$.  
For a given prompt $c=(c_1,\dots,c_M)$, the initial input state is 
\begin{equation}
x^{(T)} = [c_1, \dots, c_M, \underbrace{\texttt{[MASK]}, \dots, \texttt{[MASK]}}_{L}]
\end{equation}
where L is the pre-specified target response length.

Unlike ARMs, which extend the sequence token by token, diffusion language models (dLLMs) refine the entire sequence in parallel at every step. This sequence includes the prompt, already decoded tokens, and remaining \texttt{[MASK]} tokens. \\
At each step t, the model $f_\phi$ takes the current state $x 
^{(t)}$ and predicts a probability distribution over the vocabulary for every masked position, resulting in $y = \{y_i \mid i \in M^{(t)}\}$, the set of candidate tokens for all masked locations:
\begin{equation}
P_\phi(y \mid x^{(t)}) = f_\phi(x^{(t)})
\end{equation}
and remasking policy $\mathcal{R}$ decides which tokens to decode and which to keep masked in the next step.
\begin{equation}
x^{(t-1)} = \mathcal{R}\big(x^{(t)}, P_\phi(y \mid x^{(t)})\big)
\end{equation}
The process continues until step t=0, at which point all \texttt{[MASK]} tokens have been replaced to yield the final sequence. 
This design enables parallel multi-token prediction but at the expense of substantial computational and memory costs. 

\noindent \textbf{Caching in Diffusion Large Language Models }
Given the initial input $x^{(T)}$ composed of a prompt of length $M$ and $L$ masked positions, we denote the prompt and response token sets as $\mathcal{P}=\{1\!:\!M\}$ and $\mathcal{R}=\{M\!+\!1\!:\!M\!+\!L\}$, respectively. We index layers by $\ell=1,\ldots,D$. At denoising step $t$, for layer $\ell$ and position $i$, we define the cacheable feature bundle 
\begin{equation}
\mathrm{Feat}_{\ell}^{(t)}(i)
= \{ K_{\ell,i}^{(t)}, V_{\ell,i}^{(t)}, \mathrm{Attn}_{\ell,i}^{(t)}, \mathrm{FFN}_{\ell,i}^{(t)} \}
\end{equation}
and denote the cache entry by $\mathcal{C}_{\ell}^{(t)}(i)$. And each layer forward pass computes:
\begin{equation}
  h_{\ell}^{(t)} 
  = \mathrm{Attn}_{\ell}\!\bigl(Q_{\ell}^{(t)}, K_{\ell}^{(t)}, V_{\ell}^{(t)}\bigr)
  + \mathrm{FFN}_{\ell}
  \label{eq:layer_forward}
\end{equation}
Unlike autoregressive models, where causal attention allows past KV states to be directly reused, diffusion LLMs employ bidirectional attention over both prompt and masked positions, which prevents the reuse of intermediate states. Because this operation is repeated for every step and every token, it introduces substantial computational redundancy. Empirically, however,the feature bundles $\mathrm{Feat}_{\ell}^{(t)}(i)$ exhibit strong similarity across adjacent steps, suggesting opportunities for caching~\citep{liu2025dllm,wu2025fast,ma2025dkv,hu2025accelerating}.

At each step $t$, if $i \in \mathcal{S}^{(t)}$, we refresh 
$\mathcal{C}_{\ell}^{(t)}(i)=\mathrm{Feat}_{\ell}^{(t)}(i)$;  
otherwise, we reuse $\mathcal{C}_{\ell}^{(t-1)}(i)$. The definition of $\mathcal{S}^{(t)}$ differs by token type.  

\paragraph{Prompt tokens.}
For prompt tokens ($i\in\mathcal{P}$), we refresh them only every $T_{p}$ steps, 
i.e., $\mathcal{S}_{\mathcal{P}}^{(t)}=\{\,i\in\mathcal{P}\mid t\bmod T_{p}=0\,\}$; 
between refreshes, their cached features remain fixed.

\paragraph{Response tokens.}
For response tokens, we define two candidate refresh sets: 
$\mathcal{S}_{\mathrm{period}}^{(t)}=\{\,i\in\mathcal{R}\mid t\bmod T_r=0\,\}$ 
representing tokens refreshed periodically every $T_r$ steps, and 
$\mathcal{S}_{\mathrm{shift}}^{(t)}=\{\,i\in\mathcal{R}\mid 
\cos(V_{\ell,i}^{(t)},V_{\ell,i}^{(t-1)})<\delta\,\}$ 
representing tokens whose feature vectors vary significantly between consecutive steps.
The overall refresh set is given by 
$\mathcal{S}_{\mathcal{R}}^{(t)}=
\mathcal{S}_{\mathrm{period}}^{(t)}\cup
\mathcal{S}_{\mathrm{shift}}^{(t)}$.

\subsection{Observations}
Our hierarchical compression strategy stems from a top-down analysis of the dLLM's attention architecture, which revealed optimization opportunities at the layer, head, and token levels.

\paragraph{Layer-Level.}
Our analysis begins at the highest architectural level: the importance of each Transformer layer. We quantify this importance by measuring each layer's representational transformation, which we formalize as an \textbf{importance score} (see Section~\ref{sec: method}). Our analysis of these scores reveals two critical phenomena, both visualized in Fig.~\ref{fig:attention_map}~(b): a distinct \textbf{bimodal importance profile} and strong \textbf{cross-sample consistency}.

The bimodal profile is visually evident as the boundary layers (the first and last columns) consistently exhibit lighter colors than the middle layers, signifying lower cosine similarity and thus higher importance. The cross-sample consistency is demonstrated by the vertical uniformity of this pattern across all samples, indicating that this importance hierarchy is remarkably stable.
These two findings form the cornerstone of our allocation strategy. The bimodal profile dictates the need for a non-uniform, group-based allocation, while the cross-sample consistency validates the feasibility of a static, offline approach.

\paragraph{Head-Level.}
Building on our layer-level findings, we next analyzed the behavior of individual attention heads. Our analysis reveals a significant functional heterogeneity, as different heads exhibit widely varying degrees of dependency on the prompt context. The examples in Fig.~\ref{fig:attention_map}~(c) clearly illustrate this. The head we term the \textbf{Information Head} is highly dependent on the prompt to perform long-range information retrieval. Conversely, the \textbf{Structure Head} shows a low dependency on the prompt, as its primary role is to plan the discourse structure by organizing the syntactic framework. This observed variation is the direct motivation for our next strategy: a fine-grained, head-wise budget allocation, weighted according to each head's reliance on the prompt context.

\paragraph{Token-Level.}
Finally, after allocating budgets to layers and heads, a universal criterion is needed to select the specific tokens to retain. Our analysis of token-level attention patterns reveals this criterion. A crucial dichotomy emerges (see Fig.~\ref{fig:attention_map} (a)): non-mask tokens (\textit{i.e.}\, the prompt and already-decoded tokens) exhibit a strong \textbf{locality bias}, serving to encode their own context. In stark contrast, attention from mask queries is \textbf{highly sparse and long-range}, functioning as a task-driven information retrieval mechanism. Critically, this sparse attention is also \textbf{remarkably consistent} across all generation steps. This provides the final piece of our framework: the attention from mask queries serves as the definitive and stable signal for token importance, applicable to any head.

\subsection{The \emph{MaskKV} Framework}
\label{sec: method}
Based on our empirical findings, we propose \emph{MaskKV}, a framework that reframes KV cache pruning as a two-stage process: first, it establishes a universal importance ranking for all tokens, and second, it applies a hierarchical budget to this ranking to perform eviction.

\paragraph{Stage 1: Universal Token Importance Ranking.}

We observe that, unlike prompt queries which exhibit local bias, mask queries perform sparse long‑range retrieval with strong cross‑step consistency, making them reliable indicators of token importance. This aligns with our theoretical proof (Appendix~\ref{the primacy of mask attention}) that mask‑guided attention provides the most direct signal for the generative task.
Consequently, we propose \textbf{Mask-Voting}, a one-shot method that leverages the reliable, task-aligned signal from mask queries at the initial inference step.

We first compute the attention score matrix $A$ from all mask queries ($Q_{mask}$) to all keys ($K_{full}$):
\begin{equation}
    A = \text{softmax}_{\text{row}}\left(\frac{Q_{mask} K_{full}^T}{\sqrt{d_k}}\right)
\end{equation}
From this matrix, we derive an importance score vector $I \in \mathbb{R}^{n_p}$ for the prompt tokens by aggregating the scores each key receives:
\begin{equation}
    I_j = \sum_{i=1}^{n_m} A_{ij} \quad \text{for } j \in \{1, \dots, n_p\}
\end{equation}
The output of this stage is the vector $I$, which provides a universal, budget-agnostic importance ranking of all prompt tokens.

\paragraph{Stage 2: Hierarchical Budgeted Eviction.}
The second stage determines the fine-grained, per-head eviction budget $k_{l,h}$ using a top-down allocation policy. This process systematically distributes the total prompt budget, $k_p$, across the model's layers and then its heads, guided by data-driven importance metrics.

At the \textbf{layer level}, we allocate the total budget $k_p$ based on layer importance. We quantify this importance using a score, $I^{(l)}$, that measures the representational transformation performed by each layer. A larger transformation signifies higher importance. Inspired by prior work~\citep{wang2024squeezeattention, he2024matters}, we define it as:
\begin{equation}
\label{eq:layer_importance_metric}
    I^{(l)} = 1 - \frac{1}{n} \sum_{i=1}^{n} \text{cos\_sim}\left(h_{in, i}^{(l)}, h_{out, i}^{(l)}\right)
\end{equation}
where $h_{in, i}^{(l)}$ and $h_{out, i}^{(l)}$ are the input and output hidden states of the attention sub-layer for the $i$-th token.

With the layer importance score $I^{(l)}$ defined, we now detail our hybrid allocation strategy. The total budget $k_p$ is first partitioned into a uniform base component and an importance-driven component. The per-layer base budget, $k_{\text{base}}$, is set by a hyperparameter $\beta \in [0,1]$:
\begin{align}
    k_{\text{base}} &= \left\lfloor \frac{\beta \cdot k_p}{L} \right\rfloor \\
    k_{\text{imp}} &= k_p - L \cdot k_{\text{base}}
\end{align}
where $k_{\text{imp}}$ is the total remaining budget to be allocated based on importance. 

Next, we distribute $k_{\text{imp}}$ to the boundary ($S_{\text{bound}}$) and middle ($S_{\text{mid}}$) layer groups, proportional to their aggregated importance scores ($I_g = \sum_{l \in S_g} I^{(l)}$). This yields the \textit{total group budgets}, denoted $k_{\text{group},B}$ and $k_{\text{group},M}$:
\begin{align}
\label{eq:group_budget_allocation}
k_{\text{group},B} &= k_{\text{imp}} \cdot \frac{I_B}{I_B + I_M} \\
k_{\text{group},M} &= k_{\text{imp}} - k_{\text{group},B}
\end{align}
Finally, the budget for any given layer, $k_l$, is the sum of its base budget and an equal share of its group's importance-driven budget:
\begin{equation}
\label{eq:final_layer_budget}
    k_l = k_{\text{base}} + 
    \begin{cases} 
        \left\lfloor \dfrac{k_{\text{group},B}}{|S_{\text{bound}}|} \right\rfloor & \text{if } l \in S_{\text{bound}} \\
        \left\lfloor \dfrac{k_{\text{group},M}}{|S_{\text{mid}}|} \right\rfloor & \text{if } l \in S_{\text{mid}} 
    \end{cases}
\end{equation}

An ``online'' implementation of this algorithm is fundamentally flawed for reducing peak memory. Budget allocation requires scores from all layers, which necessitates a full forward pass with the entire, uncompressed KV cache stored. Eviction could only occur \textit{after} the peak memory has already been reached. Therefore, leveraging our finding of \textbf{cross-sample consistency}, we adopt an \textbf{offline} paradigm. By pre-computing a importance profile on a calibration set, we can determine all budgets \textit{a priori}, enabling immediate layer-by-layer eviction that effectively reduces the peak memory footprint.

At the \textbf{head level}, the per-layer budget $k_l$ is distributed among its $N_h$ heads. This allocation is guided by each head's \textbf{Prompt Preference}, $P_h^{(l)}$, which quantifies its dependency on the prompt:
\begin{equation}
\label{eq:prompt_preference_metric}
    P_h^{(l)} = \frac{S_{m \to p}^{(l,h)}}{S_{m \to p}^{(l,h)} + S_{m \to m}^{(l,h)}}
\end{equation}
where $S_{m \to p}^{(l,h)}$ and $S_{m \to m}^{(l,h)}$ are the sums of mask-to-prompt and mask-to-mask attention, respectively.

The final, fine-grained budget for each head, $k_{l,h}$, is calculated via a hybrid strategy controlled by a hyperparameter $\alpha \in [0,1]$. This strategy combines a fixed base allocation with a proportional allocation based on preference into a single formula:
\begin{equation}
\label{eq:final_head_budget}
    k_{l,h} = \left\lfloor \alpha \cdot k_l + (1-\alpha) \cdot N_h \cdot k_l \cdot \hat{P}_h^{(l)} \right\rfloor
\end{equation}
where $\hat{P}_h^{(l)} = P_h^{(l)} / \sum_{j=1}^{N_h} P_j^{(l)}$ is the normalized Prompt Preference score.

Finally, with the universal importance ranking $I$ from Stage 1 and the specific per-head budget $k_{l,h}$ from Stage 2, the eviction is performed by selecting the top-k tokens:
\begin{equation}
    S_{\text{keep}} = \operatorname{arg\,topk}(I, k_{l,h})
\end{equation}

\section{Experiment}
\subsection{Experiment Settings}

\noindent \textbf{Implementation Details.}
We evaluate the effectiveness of our method on two representative dLLMs, including LLaDA-8B-Instruct~\citep{nie2025large} and Dream-7B-Instruct~\citep{ye2025dream}.
For long-context evaluation, we follow the strategy of LongLLaDA~\citep{liu2025longllada} to ensure reliable performance on extended sequences.
All experiments were conducted on 8 × NVIDIA A100 80GB GPUs. Additional details are provided in Appendix~\ref{implement details}.

\subsection{Main Results}
\begin{table*}[t!]
    \centering
    \small
    \caption{LongBench results for LLaDA-8B and Dream-7B with specific KV cache budgets (B=32 and B=128). Best result in each column within a budget section is in \textbf{bold}.}
    \label{tab:main_b32_b128_results}
    \setlength{\tabcolsep}{3pt} 
    \renewcommand{\arraystretch}{1.1}
    \resizebox{\textwidth}{!}{%
    \begin{tabular}{@{}l*{15}{c}@{}}
        \toprule
        \textbf{Method} & \multicolumn{2}{c}{Single-Doc. QA} & \multicolumn{3}{c}{Multi-Doc. QA} & \multicolumn{3}{c}{Summarization} & \multicolumn{3}{c}{Few-shot Learning} & \multicolumn{1}{c}{Synthetic} & \multicolumn{2}{c}{Code} & \makecell{Ave. \\ Score} \\
        
        \cmidrule(lr){2-3} \cmidrule(lr){4-6} \cmidrule(lr){7-9} \cmidrule(lr){10-12} \cmidrule(lr){13-13} \cmidrule(lr){14-15}

        & \rotatebox[origin=c]{-45}{Qasper} & \rotatebox[origin=c]{-45}{MF-en} & \rotatebox[origin=c]{-45}{HotpotQA} 
        & \rotatebox[origin=c]{-45}{2WikiMQA} & \rotatebox[origin=c]{-45}{Musique} 
        & \rotatebox[origin=c]{-45}{GovReport} & \rotatebox[origin=c]{-45}{QMSum} & \rotatebox[origin=c]{-45}{MultiNews}
        & \rotatebox[origin=c]{-45}{TREC} & \rotatebox[origin=c]{-45}{TriviaQA} & \rotatebox[origin=c]{-45}{SAMSum}
        & \rotatebox[origin=c]{-45}{PRe}
        & \rotatebox[origin=c]{-45}{Lcc} & \rotatebox[origin=c]{-45}{RB-P}
        & \\
        
        \midrule
        \multicolumn{16}{c}{\large \textbf{LLaDA-8B-Instruct}} \\
        \midrule
        \multicolumn{16}{c}{\small Full KV Cache} \\
        \midrule
    \textcolor{gray}{dLLM w/o Cache} & \textcolor{gray}{16.96} & \textcolor{gray}{31.31} & \textcolor{gray}{14.68} & \textcolor{gray}{17.60} & \textcolor{gray}{11.48} & \textcolor{gray}{29.24} & \textcolor{gray}{21.93} & \textcolor{gray}{27.58} & \textcolor{gray}{65.20} & \textcolor{gray}{47.98} & \textcolor{gray}{40.51} & \textcolor{gray}{98.17} & \textcolor{gray}{65.69} & \textcolor{gray}{59.57} & \textcolor{gray}{39.14} \\
    \textcolor{gray}{dLLM w/ Cache} & \textcolor{gray}{15.26} & \textcolor{gray}{29.62} & \textcolor{gray}{13.87} & \textcolor{gray}{17.17} & \textcolor{gray}{10.44} & \textcolor{gray}{29.75} & \textcolor{gray}{22.06} & \textcolor{gray}{26.68} & \textcolor{gray}{66.00} & \textcolor{gray}{44.94} & \textcolor{gray}{41.86} & \textcolor{gray}{97.44} & \textcolor{gray}{66.07} & \textcolor{gray}{59.34} & \textcolor{gray}{38.61} \\

        \midrule
        \multicolumn{16}{c}{\small B=32} \\
        \midrule
        SnapKV & 9.10 & 17.49 & \textbf{17.38} & 16.45 & 8.06 & \textbf{9.92} & 12.21 & 13.95 & 39.25 & 54.32 & 16.41 & 55.00 & 39.90 & 29.08 & 24.18 \\
        PyramidKV & 9.90 & 12.12 & 14.46 & 14.62 & 8.18 & 9.10 & 8.83 & 12.36 & 26.42 & 53.26 & 14.51 & 28.00 & 39.18 & 27.66 & 19.90 \\
        SqueezeAttention & 11.49 & 14.72 & 16.66 & 15.42 & 8.15 & 9.28 & 10.49 & 14.86 & 43.50 & 53.67 & 15.73 & 52.00 & 33.60 & 25.56 & 23.22 \\
        AdaKV & 11.15 & 16.21 & 16.69 & \textbf{16.98} & 7.48 & 9.10 & 10.44 & 14.23 & 39.17 & \textbf{55.69} & 19.00 & 59.50 & 37.21 & 26.40 & 24.23 \\
        \rowcolor{lightgreen} \textbf{MaskKV (Ours)} & \textbf{14.61} & \textbf{24.45} & 17.05 & 15.68 & \textbf{12.50} & 9.54 & \textbf{14.33} & \textbf{16.43} & \textbf{40.42} & 54.64 & \textbf{29.28} & \textbf{90.33} & \textbf{56.08} & \textbf{42.21} & \textbf{31.25} \\
        \midrule
        \multicolumn{16}{c}{\small B=128} \\
        \midrule
        SnapKV & 17.42 & 26.65 & 15.88 & 17.44 & 7.99 & 11.50 & 11.69 & 18.89 & 50.83 & 55.60 & 20.72 & 80.00 & 54.90 & 39.33 & 30.63 \\
        PyramidKV & 15.71 & 25.20 & 16.22 & 16.20 & 8.47 & 10.09 & 11.06 & 17.22 & 39.92 & 55.03 & 23.49 & 83.25 & 54.00 & 40.50 & 29.74 \\
        SqueezeAttention & 14.46 & 17.64 & \textbf{18.12} & \textbf{17.97} & 8.00 & 13.46 & 10.69 & 19.46 & 52.67 & 54.71 & 17.04 & 71.00 & 48.30 & 32.51 & 28.29 \\
        AdaKV & 19.42 & 24.68 & 17.06 & 17.57 & 8.82 & 11.85 & 9.51 & 18.49 & 49.92 & \textbf{57.90} & 21.03 & 78.00 & 54.64 & 36.27 & 30.37 \\
        \rowcolor{lightgreen} \textbf{MaskKV (Ours)} & \textbf{20.21} & \textbf{29.84} & 15.78 & 16.65 & \textbf{11.83} & \textbf{13.60} & \textbf{17.67} & \textbf{20.78} & \textbf{57.00} & 46.06 & \textbf{37.28} & \textbf{98.17} & \textbf{61.61} & \textbf{51.86} & \textbf{35.60} \\
        \midrule
        \multicolumn{16}{c}{\large \textbf{Dream-v0-Instruct-7B}} \\
        \midrule
        \multicolumn{16}{c}{\small Full KV Cache} \\
        \midrule
\textcolor{gray}{dLLM w/o Cache} & \textcolor{gray}{28.17} & \textcolor{gray}{36.23} & \textcolor{gray}{27.65} & \textcolor{gray}{32.43} & \textcolor{gray}{11.83} & \textcolor{gray}{5.04} & \textcolor{gray}{14.29} & \textcolor{gray}{5.95} & \textcolor{gray}{73.00} & \textcolor{gray}{89.25} & \textcolor{gray}{37.84} & \textcolor{gray}{16.92} & \textcolor{gray}{38.91} & \textcolor{gray}{45.08} & \textcolor{gray}{33.04} \\
\textcolor{gray}{dLLM w/ Cache} & \textcolor{gray}{26.55} & \textcolor{gray}{39.86} & \textcolor{gray}{27.66} & \textcolor{gray}{32.09} & \textcolor{gray}{11.12} & \textcolor{gray}{4.40} & \textcolor{gray}{13.89} & \textcolor{gray}{5.51} & \textcolor{gray}{73.50} & \textcolor{gray}{89.59} & \textcolor{gray}{36.07} & \textcolor{gray}{12.05} & \textcolor{gray}{39.88} & \textcolor{gray}{45.57} & \textcolor{gray}{32.70} \\

        \midrule
        \multicolumn{16}{c}{\small B=32} \\
        \midrule
        SnapKV & 17.75 & 26.82 & 22.39 & \textbf{27.91} & 7.53 & \textbf{2.65} & \textbf{12.58} & 1.95 & 28.25 & 66.14 & 23.67 & 17.50 & 22.75 & 23.87 & 21.55 \\
        PyramidKV & 14.55 & 24.51 & 22.41 & 15.27 & 6.90 & 2.62 & 11.89 & 2.15 & 28.00 & 57.55 & 24.17 & 11.50 & 22.02 & 22.22 & 18.98 \\
        SqueezeAttention & 17.62 & 30.44 & 20.07 & 26.15 & 7.42 & 2.62 & 11.83 & 2.19 & 26.25 & 72.17 & 24.52 & 17.00 & 20.91 & 23.41 & 22.00 \\
        AdaKV & 17.09 & 25.42 & 23.64 & 24.77 & 7.55 & 2.58 & 12.46 & 1.89 & 27.75 & 69.77 & 25.21 & 18.83 & 21.67 & 24.98 & 21.69 \\
        \rowcolor{lightgreen} \textbf{MaskKV (Ours)} & \textbf{18.53} & \textbf{33.76} & \textbf{32.92} & 24.53 & \textbf{11.02} & 2.59 & 11.98 & \textbf{2.25} & \textbf{33.25} & \textbf{86.47} & \textbf{27.55} & \textbf{20.00} & \textbf{24.89} & \textbf{27.40} & \textbf{25.51} \\
        \midrule
        \multicolumn{16}{c}{\small B=128} \\
        \midrule
        SnapKV & \textbf{22.36} & 39.02 & 30.28 & 32.27 & 11.51 & 2.72 & \textbf{13.20} & 2.72 & 38.00 & 87.17 & 28.69 & \textbf{26.05} & \textbf{32.55} & 36.51 & 28.79 \\
        PyramidKV & 17.31 & 36.59 & 25.21 & 24.38 & 10.99 & 2.77 & 12.75 & 2.90 & 39.75 & 84.83 & 29.71 & 24.00 & 30.26 & 30.33 & 26.56 \\
        SqueezeAttention & 19.36 & 36.19 & 28.35 & 29.79 & 10.25 & 2.81 & 13.11 & 2.64 & 33.00 & 85.99 & 29.25 & 25.00 & 27.11 & 31.92 & 26.77 \\
        AdaKV & 21.97 & \textbf{39.68} & 33.24 & 31.57 & 11.38 & 2.69 & 13.15 & 2.80 & 38.75 & 88.51 & 30.43 & 24.25 & 31.46 & \textbf{37.28} & 29.08 \\
        \rowcolor{lightgreen} \textbf{MaskKV (Ours)} & 21.49 & 39.23 & \textbf{38.25} & \textbf{33.99} & \textbf{14.76} & \textbf{2.86} & 12.48 & \textbf{3.19} & \textbf{53.50} & \textbf{88.69} & \textbf{30.97} & 21.92 & 30.71 & 34.79 & \textbf{30.49} \\
        \bottomrule
    \end{tabular}
    } 
\end{table*}
\noindent \textbf{Preserving Accuracy with Reduced Cache.}
As shown in Tab.~\ref{tab:main_b32_b128_results}, our method consistently surpasses prior cache-eviction strategies on specific budgets. Under the extreme 32 KV budget, it outperforms the best competing baseline by 7.02 points on LLaDA-8B. Notably, our method can even surpass the full-context dLLM-Cache baseline, likely because eviction removes distracting noise from the context and enhances the model's focus.

\begin{figure}[t]
  \centering
  \includegraphics[width=\columnwidth]{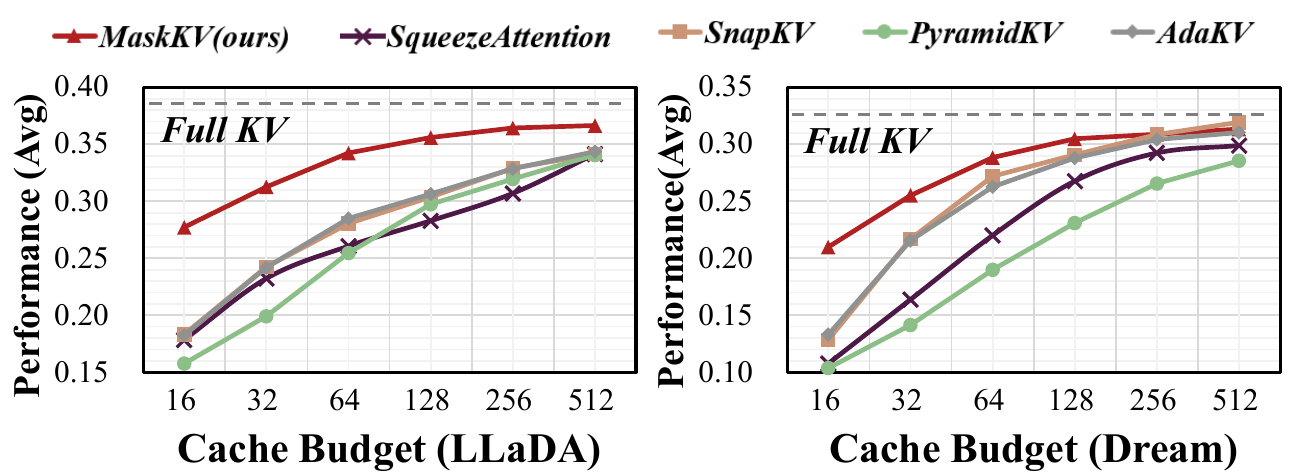}
  \caption{Average LongBench performance across varying KV cache sizes.}
  \label{fig:across_size}
\end{figure}

\noindent \textbf{Stable Performance across Varied Budgets.}
As shown in Fig.\ref{fig:across_size}, with a 256~KV budget, our method retains 94.33\% of the dLLM‑Cache baseline's performance on \textbf{LLaDA} and 98.66\% on \textbf{Dream}, achieving the best results. This advantage holds across all KV budgets and remains strong even in extremely low‑budget regimes where autoregressive models typically collapse~\citep{xiao2023efficient}. We attribute this robustness to bidirectional attention, which integrates information from the entire sequence to form richer KV representations, enabling aggressive pruning with preserving high generation quality.

\begin{figure}[t]
  \centering
  \includegraphics[width=\columnwidth]{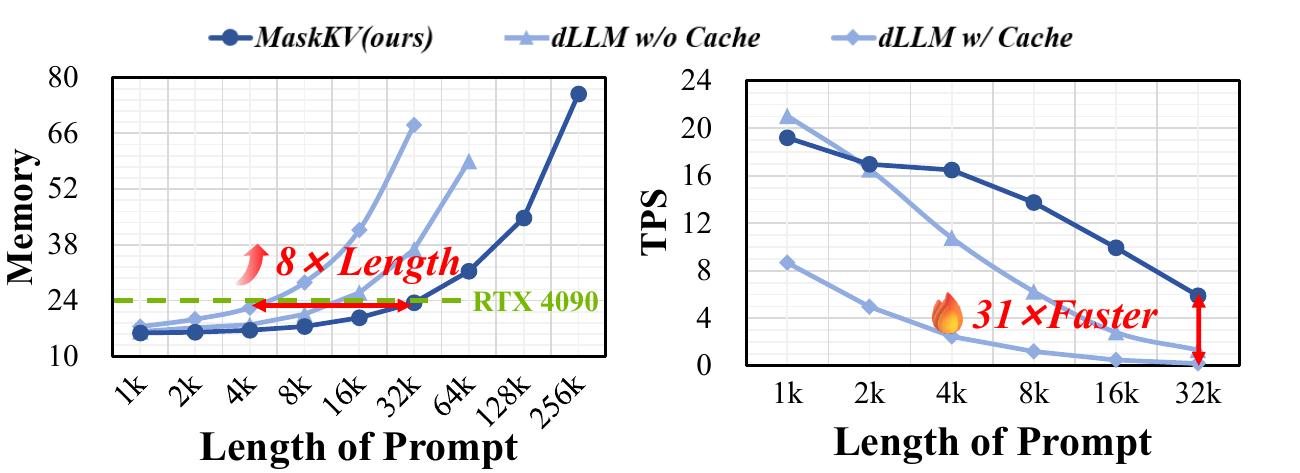}
  \caption{Analysis on latency and memory reduction.}
  \label{fig:efficiency}
\end{figure}

\noindent \textbf{Efficiency in Speed and Memory.}
Our method markedly improves throughput and memory efficiency for long-context inference. 
We introduce two implementation optimizations, \textit{Prompt-State Exclusion} and \textit{Mask-Only Projection} (details in Appendix~\ref{sec:maskkv_details}). 
With these techniques, the memory footprint of \emph{MaskKV} becomes comparable to, or even lower than, that of LLaDA under identical configurations. 
At a 32K-token context, it achieves \textbf{31$\times$} faster decoding and \textbf{65\%} lower peak memory than LLaDA, supporting up to \textbf{8$\times$} longer prompts on an RTX 4090 GPU. 
Ablation results isolating the effects of cache eviction and the proposed optimizations are provided in Tab.~\ref{tab:ablation_memory}.

\subsection{Ablation Study} 
\noindent \textbf{Effect of Base Rates $\alpha$ and $\beta$.}
\begin{figure}[t]
  \centering
  \includegraphics[width=\columnwidth]{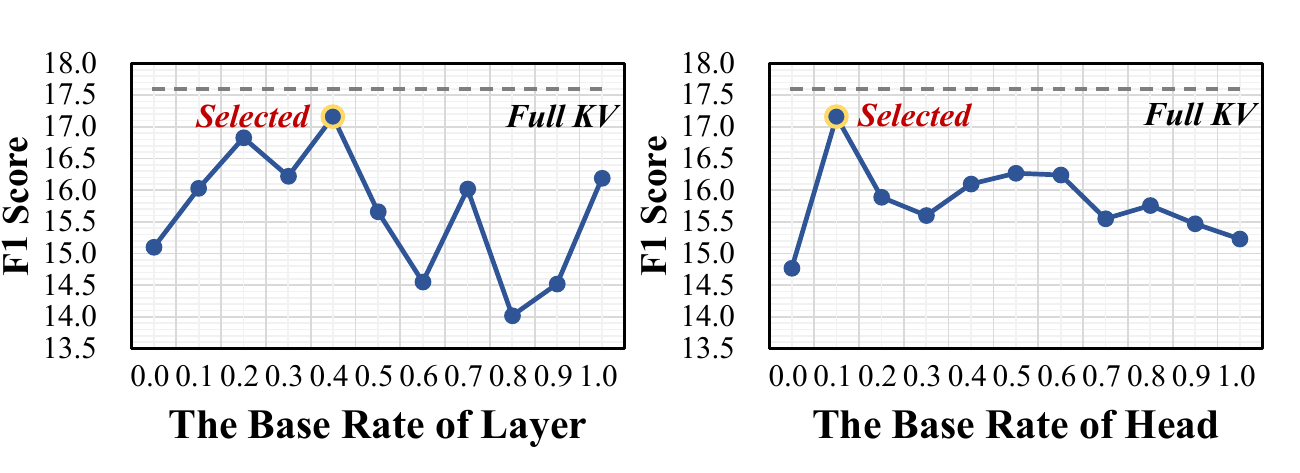}
  \caption{Impact of base rate on model performance.}
  \label{fig:hy}
\end{figure}
The base rate of head ($\alpha$) and layer ($\beta$) represent uniform budget floors for attention heads and network layers, respectively.
These base rates first guarantee each unit a minimal share, after which the remaining budget is redistributed according to estimated importance.
Excessively large values drive allocations toward near‑uniformity, diluting capacity for critical modules, whereas overly small values make the policy too aggressive and unstable.
Empirical results (Fig.~\ref{fig:hy}) show that setting $\alpha=0.1$ 
and $\beta=0.4$ provides the most stable accuracy and key-value (KV) budgets, while 
still concentrating resources where they matter most.

\begin{table}[t]
\centering
\small
\renewcommand{\arraystretch}{1.15}
\setlength{\tabcolsep}{10pt}
\begin{tabular}{l l}
\toprule
\textbf{Variant} & \textbf{Performance (Avg) } \\
\midrule
Mask-Voting  & $35.39$ \\
\midrule
\multicolumn{2}{c}{\textit{Layer Budget Allocation}} \\
\midrule
+SqueezeAttention          & $35.62_{\textcolor{green}{\scriptsize +0.23}}$ \\
+PyramidKV          & $33.52_{\textcolor{black!50}{\scriptsize -1.87}}$ \\
\rowcolor{lightgreen}
+Boundary-Aware(online)  & $\mathbf{35.65}_{\textcolor{green}{\scriptsize +0.26}}$ \\
\rowcolor{lightgreen}
+Boundary-Aware(offline)  & $\mathbf{35.74}_{\textcolor{green}{\scriptsize +0.35}}$ \\
\midrule
\multicolumn{2}{c}{\textit{Head Budget Allocation}} \\
\midrule
+adaKV         & $35.70_{\textcolor{green}{\scriptsize +0.32}}$ \\
\rowcolor{lightgreen}
+Prompt-Preference    & $\mathbf{35.96}_{\textcolor{green}{\scriptsize +0.57}}$ \\
\midrule
\multicolumn{2}{c}{\textit{Layer + Head Allocation}} \\
\midrule
\rowcolor{lightgreen}
\textbf{MaskKV (Ours) } & $\mathbf{36.27}_{\textcolor{green}{\scriptsize +0.88}}$ \\
\bottomrule
\end{tabular}
\caption{Ablation study of budget allocation.}
\label{tab:ablation}
\end{table}

\noindent \textbf{Mask-Token Voting and Budget Allocation.}
Our Mask-Voting consistently achieves superior performance over other token selection methods by directly leveraging mask queries, which helps avoid local bias and better identify influential tokens. More comprehensive comparisons with alternative token selection strategies are provided in  Tab.~\ref{tab:voting}. In addition, our budget allocation strategies Boundary-Aware at the layer level and Prompt-Preference at the head level both outperform competing approaches in Tab.~\ref{tab:ablation}.

\section{Discussion}
\noindent \textbf{Needle‑in‑a‑Haystack.}
\begin{figure}[t]
  \centering
  \includegraphics[width=\columnwidth]{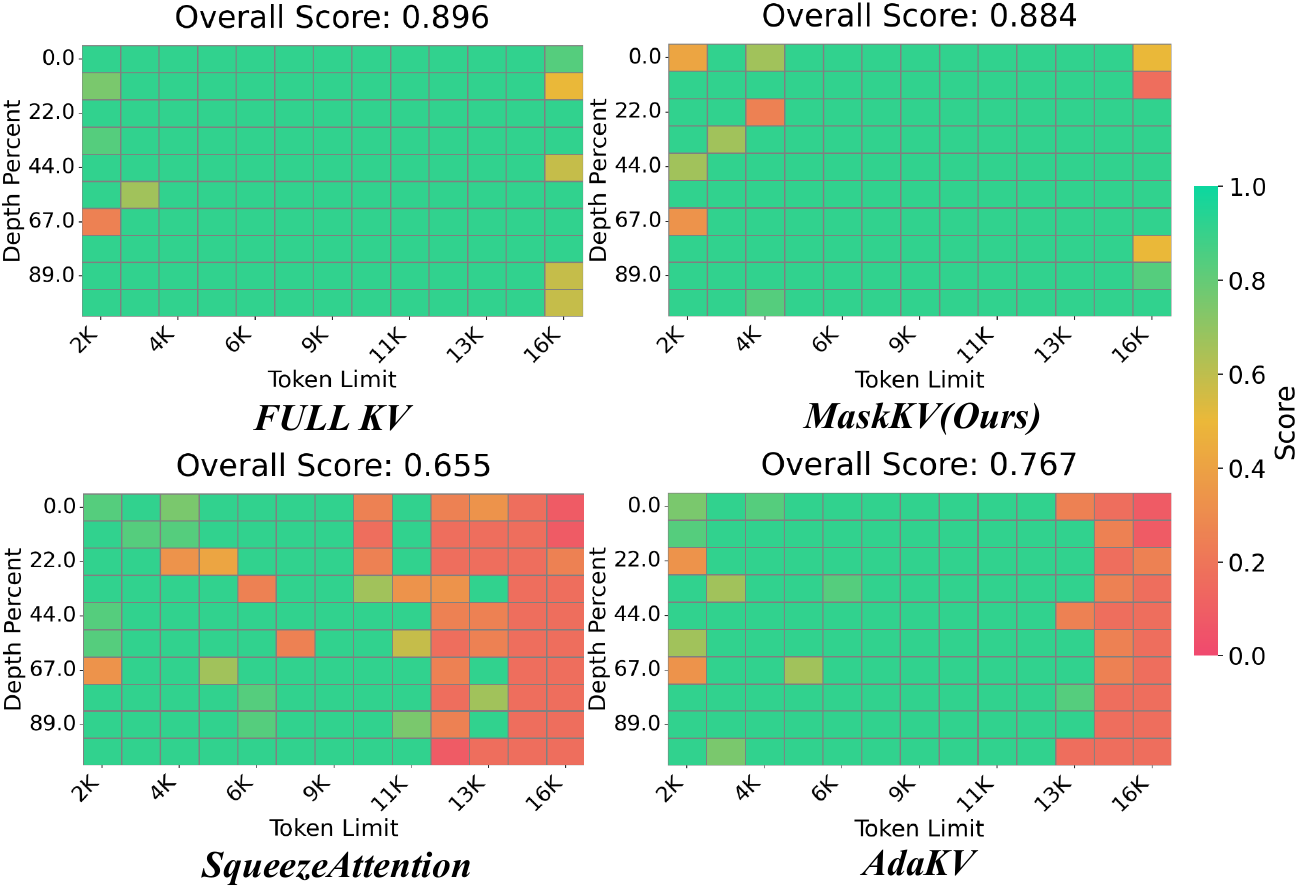}
    \caption{A visualization from the Needle-in-a-Haystack test. See Figure~\ref{fig:niah_all} for full results.}
  \label{fig:needle}
\end{figure}
To better understand the model’s ability to retrieve fine‑grained information hidden within long contexts, we conduct a preliminary study using the “needle‑in‑a‑haystack” setup (see Fig.~\ref{fig:needle}). Our method shows stronger robustness to increasing prompt length, effectively maintaining information retrieval performance.

\noindent \textbf{Visualization of Prompt Preference.}

\begin{figure}[H]
  \centering
  \includegraphics[width=\columnwidth]{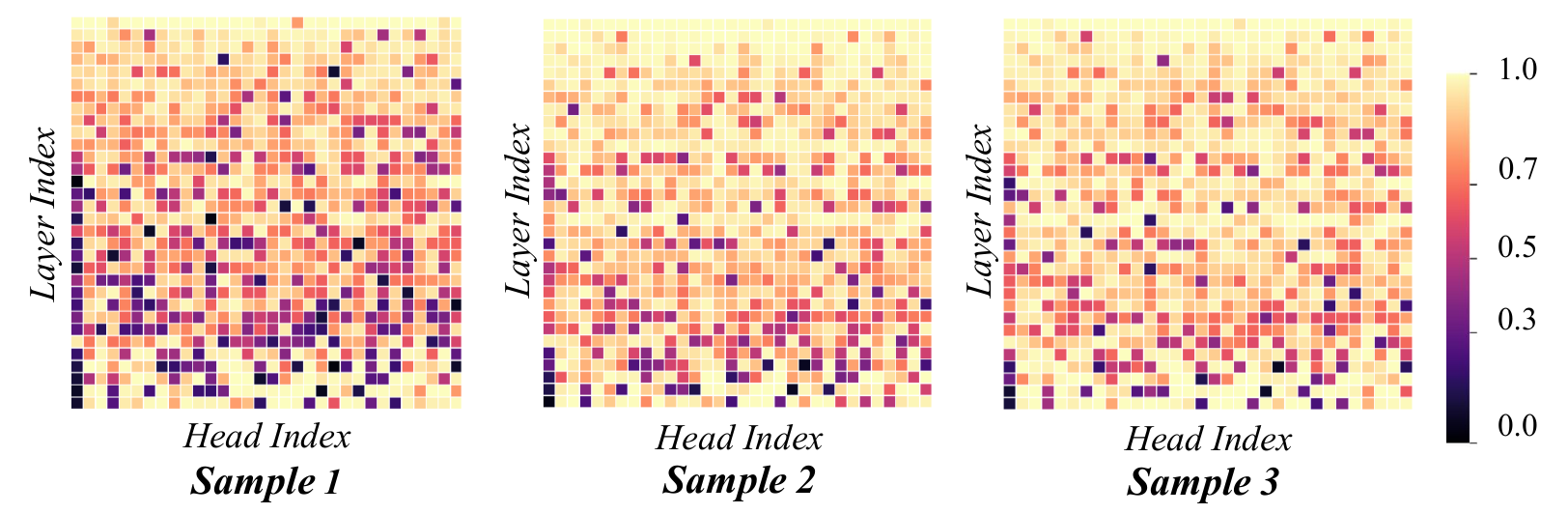}
  \caption{Visualization of Prompt Preference.}
  \label{fig:prefer}
\end{figure}

As shown in Fig.~\ref{fig:prefer}, we analyze the prompt preference distribution across different heads within the same layer. Some heads allocate substantial attention to the prompt, likely to extract task-relevant information, while others focus more on the masked region to plan answer generation. Such analysis provides deeper insight into their internal decision behavior.

\noindent \textbf{ Comparison of Voting Strategies.}
We analyze voting strategies by grouping masked tokens into front, middle, and back regions, each contributing a position-specific vote. Results (Tab.~\ref{tab:mask_voting_position}) show that later positions yield more accurate votes for identifying critical tokens, suggesting position-aware voting may further improve eviction effectiveness.

\section{Conclusion}
We explored dLLM characteristics and introduced \mymethod, a training‑free framework enabling fine‑grained cache eviction via Mask Voting and adaptive layer–head budget allocation. Experiments show that \mymethod~reduces the KV cache to 256 tokens while retaining up to 94\% of original performance, highlighting an efficient trade‑off for long‑context inference.

\section{Limitations}
Our current experiments are limited to 7B/8B-scale dLLMs, constrained by the limited availability of open-source models.
The effectiveness of our proposed methods and the observed attention behavior patterns have yet to be validated on both larger-scale and smaller lightweight models.
Moreover, our evaluation is confined to text generation benchmarks, and extending the analysis to multimodal reasoning remains an important direction for future research.
\newpage
\newpage
\bibliography{custom}

\appendix
\section{Appendix}
\label{sec:appendix}
\begin{figure*}
  \centering
  \includegraphics[width=1.0\linewidth]{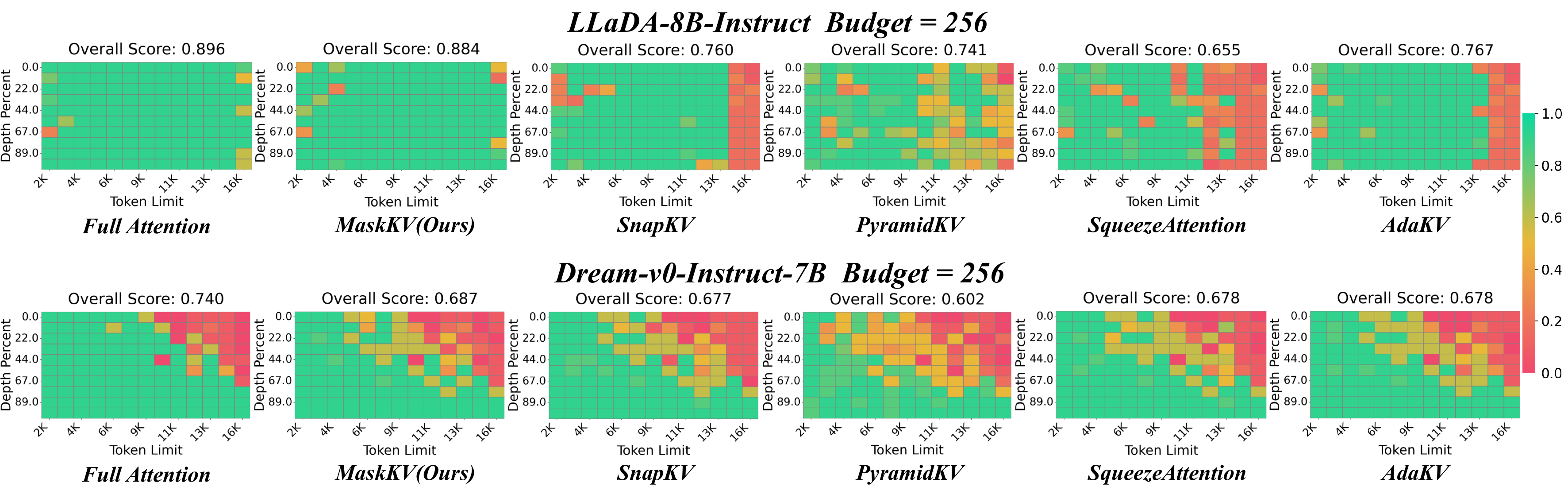}
  \caption{Performance comparison of different KV Cache compression techniques on LLaDA-8B and Dream-7B models in the ``Needle-in-a-Haystack'' test (Budget B=256). The heatmaps show retrieval accuracy at different context lengths (x-axis) and depths (y-axis), where greener colors indicate better performance.}
  \label{fig:niah_all}
\end{figure*}

\begin{table*}[t!]
    \centering
    \small
    \caption{Detailed LongBench Results for LLaDA-8B-Instruct. Best result in each column within a budget section is in \textbf{bold}.}
    \label{tab:llada_full_longbench_results_final_reordered}
    
    \setlength{\tabcolsep}{3pt} 
    \renewcommand{\arraystretch}{1.1}
    
    \resizebox{\textwidth}{!}{%
    \begin{tabular}{@{}l*{15}{c}@{}}
        \toprule
        \textbf{Method} & \multicolumn{2}{c}{Single-Doc. QA} & \multicolumn{3}{c}{Multi-Doc. QA} & \multicolumn{3}{c}{Summarization} & \multicolumn{3}{c}{Few-shot Learning} & \multicolumn{1}{c}{Synthetic} & \multicolumn{2}{c}{Code} & \makecell{Ave. \\ Score} \\
        
        \cmidrule(lr){2-3} \cmidrule(lr){4-6} \cmidrule(lr){7-9} \cmidrule(lr){10-12} \cmidrule(lr){13-13} \cmidrule(lr){14-15}

        & \rotatebox[origin=c]{-45}{Qasper} & \rotatebox[origin=c]{-45}{MF-en} & \rotatebox[origin=c]{-45}{HotpotQA} 
        & \rotatebox[origin=c]{-45}{2WikiMQA} & \rotatebox[origin=c]{-45}{Musique} 
        & \rotatebox[origin=c]{-45}{GovReport} & \rotatebox[origin=c]{-45}{QMSum} & \rotatebox[origin=c]{-45}{MultiNews}
        & \rotatebox[origin=c]{-45}{TREC} & \rotatebox[origin=c]{-45}{TriviaQA} & \rotatebox[origin=c]{-45}{SAMSum}
        & \rotatebox[origin=c]{-45}{PRe}
        & \rotatebox[origin=c]{-45}{Lcc} & \rotatebox[origin=c]{-45}{RB-P}
        & \\
        
        \midrule
        \multicolumn{16}{c}{\large \textbf{LLaDA-8B-Instruct}} \\
        \midrule
        \multicolumn{16}{c}{\small Full KV Cache} \\
        \midrule
        dLLM w/o Cache & 16.96 & 31.31 & 14.68 & 17.60 & 11.48 & 29.24 & 21.93 & 27.58 & 65.20 & 47.98 & 40.51 & 98.17 & 65.69 & 59.57 & 39.14 \\
        dLLM w/ Cache & 15.26 & 29.62 & 13.87 & 17.17 & 10.44 & 29.75 & 22.06 & 26.68 & 66.00 & 44.94 & 41.86 & 97.44 & 66.07 & 59.34 & 38.61 \\
        \midrule
        \multicolumn{16}{c}{\small B=16} \\
        \midrule
        SnapKV & 10.66 & 12.28 & 12.96 & 13.18 & 4.91 & 7.59 & 9.85 & 11.80 & 21.25 & 49.98 & 14.37 & 31.50 & 30.75 & 24.17 & 18.23 \\
        PyramidKV & 7.94 & 8.34 & 10.95 & 8.77 & 4.80 & 7.61 & 9.02 & 10.98 & 14.75 & 46.29 & 12.71 & 24.00 & 30.48 & 23.94 & 15.76 \\
        SqueezeAttention & 8.27 & 10.49 & 14.28 & 14.19 & 5.47 & 7.27 & 7.92 & 11.98 & 31.25 & 42.39 & 13.83 & 30.25 & 29.49 & 22.73 & 17.84 \\
        AdaKV & 9.48 & 11.97 & 12.13 & \textbf{15.16} & 5.62 & 7.55 & \textbf{11.34} & 11.90 & 25.00 & 49.16 & 14.24 & 31.58 & 28.63 & 22.81 & 18.33 \\
        \rowcolor{lightgreen} \textbf{MaskKV (Ours)} & \textbf{16.11} & \textbf{19.81} & \textbf{18.24} & 14.16 & \textbf{11.40} & \textbf{7.86} & 8.31 & \textbf{13.69} & \textbf{28.50} & \textbf{52.32} & \textbf{23.29} & \textbf{87.00} & \textbf{50.09} & \textbf{37.32} & \textbf{27.72} \\
        \midrule
        \multicolumn{16}{c}{\small B=32} \\
        \midrule
        SnapKV & 9.10 & 17.49 & \textbf{17.38} & 16.45 & 8.06 & \textbf{9.92} & 12.21 & 13.95 & 39.25 & 54.32 & 16.41 & 55.00 & 39.90 & 29.08 & 24.18 \\
        PyramidKV & 9.90 & 12.12 & 14.46 & 14.62 & 8.18 & 9.10 & 8.83 & 12.36 & 26.42 & 53.26 & 14.51 & 28.00 & 39.18 & 27.66 & 19.90 \\
        SqueezeAttention & 11.49 & 14.72 & 16.66 & 15.42 & 8.15 & 9.28 & 10.49 & 14.86 & 43.50 & 53.67 & 15.73 & 52.00 & 33.60 & 25.56 & 23.22 \\
        AdaKV & 11.15 & 16.21 & 16.69 & \textbf{16.98} & 7.48 & 9.10 & 10.44 & 14.23 & 39.17 & \textbf{55.69} & 19.00 & 59.50 & 37.21 & 26.40 & 24.23 \\
        \rowcolor{lightgreen} \textbf{MaskKV (Ours)} & \textbf{14.61} & \textbf{24.45} & 17.05 & 15.68 & \textbf{12.50} & 9.54 & \textbf{14.33} & \textbf{16.43} & \textbf{40.42} & 54.64 & \textbf{29.28} & \textbf{90.33} & \textbf{56.08} & \textbf{42.21} & \textbf{31.25} \\
        \midrule
        \multicolumn{16}{c}{\small B=64} \\
        \midrule
        SnapKV & 13.77 & 23.25 & 17.12 & \textbf{18.74} & 8.74 & 10.55 & 10.78 & 16.67 & 46.83 & 56.95 & 19.79 & 73.00 & 48.94 & 33.75 & 28.49 \\
        PyramidKV & 12.23 & 21.13 & 15.95 & 16.90 & 8.64 & 9.42 & 9.24 & 15.02 & 37.25 & 56.70 & 18.70 & 51.00 & 49.73 & 34.53 & 25.46 \\
        SqueezeAttention & 11.30 & 17.71 & 17.68 & 18.32 & 9.49 & 11.20 & 11.60 & 17.61 & 44.79 & 54.58 & 16.51 & 66.00 & 38.64 & 29.51 & 26.07 \\
        AdaKV & 13.21 & 22.39 & \textbf{19.60} & 17.56 & 8.25 & 10.50 & 9.25 & 17.25 & 47.17 & \textbf{57.65} & 20.15 & 71.50 & 47.88 & 30.44 & 28.06 \\
        \rowcolor{lightgreen} \textbf{MaskKV (Ours)} & \textbf{18.48} & \textbf{27.42} & 19.00 & 16.04 & \textbf{10.12} & \textbf{10.93} & \textbf{17.25} & \textbf{18.88} & \textbf{55.08} & 50.98 & \textbf{33.47} & \textbf{95.08} & \textbf{59.71} & \textbf{46.80} & \textbf{34.23} \\
        \midrule
        \multicolumn{16}{c}{\small B=128} \\
        \midrule
        SnapKV & 17.42 & 26.65 & 15.88 & 17.44 & 7.99 & 11.50 & 11.69 & 18.89 & 50.83 & 55.60 & 20.72 & 80.00 & 54.90 & 39.33 & 30.63 \\
        PyramidKV & 15.71 & 25.20 & 16.22 & 16.20 & 8.47 & 10.09 & 11.06 & 17.22 & 39.92 & 55.03 & 23.49 & 83.25 & 54.00 & 40.50 & 29.74 \\
        SqueezeAttention & 14.46 & 17.64 & \textbf{18.12} & \textbf{17.97} & 8.00 & 13.46 & 10.69 & 19.46 & 52.67 & 54.71 & 17.04 & 71.00 & 48.30 & 32.51 & 28.29 \\
        AdaKV & 19.42 & 24.68 & 17.06 & 17.57 & 8.82 & 11.85 & 9.51 & 18.49 & 49.92 & \textbf{57.90} & 21.03 & 78.00 & 54.64 & 36.27 & 30.37 \\
        \rowcolor{lightgreen} \textbf{MaskKV (Ours)} & \textbf{20.21} & \textbf{29.84} & 15.78 & 16.65 & \textbf{11.83} & \textbf{13.60} & \textbf{17.67} & \textbf{20.78} & \textbf{57.00} & 46.06 & \textbf{37.28} & \textbf{98.17} & \textbf{61.61} & \textbf{51.86} & \textbf{35.60} \\
        \midrule
        \multicolumn{16}{c}{\small B=256} \\
        \midrule
        SnapKV & 19.74 & 27.33 & 16.62 & \textbf{18.41} & \textbf{10.35} & 12.12 & 12.03 & 21.00 & 47.63 & 53.29 & 26.53 & 92.92 & 58.93 & 42.95 & 32.85 \\
        PyramidKV & 18.50 & 29.07 & 14.61 & 16.05 & 8.92 & 12.65 & 12.15 & 19.22 & 43.17 & 52.17 & 29.09 & 91.33 & 56.54 & 43.89 & 31.95 \\
        SqueezeAttention & 19.35 & 23.79 & 16.69 & 16.35 & 8.78 & 13.75 & 11.24 & 21.43 & 51.50 & \textbf{54.12} & 21.48 & 77.25 & 57.08 & 36.91 & 30.69 \\
        AdaKV & 19.52 & 27.73 & 16.14 & 17.36 & 9.89 & 12.12 & 11.41 & 20.13 & 51.25 & 53.88 & 25.87 & 96.25 & 58.30 & 40.78 & 32.90 \\
        \rowcolor{lightgreen} \textbf{MaskKV (Ours)} & \textbf{22.71} & \textbf{29.72} & \textbf{14.81} & 16.77 & 9.19 & \textbf{16.69} & \textbf{20.18} & \textbf{23.14} & \textbf{62.25} & 41.65 & \textbf{40.48} & \textbf{98.92} & \textbf{61.23} & \textbf{52.20} & \textbf{36.42} \\
        \midrule
        \multicolumn{16}{c}{\small B=512} \\
        \midrule
        SnapKV & 19.98 & 30.03 & 12.89 & 16.93 & \textbf{10.32} & 15.66 & 13.21 & 23.36 & 49.17 & 52.09 & 32.71 & 98.00 & 61.24 & 45.37 & 34.35 \\
        PyramidKV & 19.53 & 29.23 & 14.06 & 14.75 & 9.31 & 15.45 & 13.93 & 21.81 & 52.67 & 46.03 & 35.28 & \textbf{100.00} & 59.28 & 45.58 & 34.07 \\
        SqueezeAttention & \textbf{20.71} & 26.45 & \textbf{16.44} & \textbf{17.70} & 8.04 & 14.69 & 12.23 & 23.54 & 55.92 & 50.13 & 29.29 & 96.75 & 61.08 & 45.01 & 34.14 \\
        AdaKV & 19.65 & \textbf{31.77} & 13.47 & 16.10 & 9.97 & 14.53 & 13.41 & 22.34 & 50.25 & 52.06 & 33.31 & 96.00 & 60.95 & 44.27 & 34.15 \\
        \rowcolor{lightgreen} \textbf{MaskKV (Ours)} & 17.85 & 28.92 & 13.85 & 17.10 & 8.93 & \textbf{18.58} & \textbf{20.29} & \textbf{25.05} & \textbf{64.25} & \textbf{41.62} & \textbf{41.04} & 99.33 & \textbf{62.46} & \textbf{53.78} & \textbf{36.65} \\

        \bottomrule
    \end{tabular}
    } 
\end{table*}
\begin{table*}[t!]
    \centering
    \small
    \caption{Detailed LongBench Results for Dream-v0-Instruct-7B. Best result in each column within a budget section is in \textbf{bold}.}
    \label{tab:dream_full_longbench_results_final_verified}
    
    \setlength{\tabcolsep}{3pt} 
    \renewcommand{\arraystretch}{1.1}
    
    \resizebox{\textwidth}{!}{%
    \begin{tabular}{@{}l*{15}{c}@{}}
        \toprule
        \textbf{Method} & \multicolumn{2}{c}{Single-Doc. QA} & \multicolumn{3}{c}{Multi-Doc. QA} & \multicolumn{3}{c}{Summarization} & \multicolumn{3}{c}{Few-shot Learning} & \multicolumn{1}{c}{Synthetic} & \multicolumn{2}{c}{Code} & \makecell{Ave. \\ Score} \\
        
        \cmidrule(lr){2-3} \cmidrule(lr){4-6} \cmidrule(lr){7-9} \cmidrule(lr){10-12} \cmidrule(lr){13-13} \cmidrule(lr){14-15}

        & \rotatebox[origin=c]{-45}{Qasper} & \rotatebox[origin=c]{-45}{MF-en} & \rotatebox[origin=c]{-45}{HotpotQA} 
        & \rotatebox[origin=c]{-45}{2WikiMQA} & \rotatebox[origin=c]{-45}{Musique} 
        & \rotatebox[origin=c]{-45}{GovReport} & \rotatebox[origin=c]{-45}{QMSum} & \rotatebox[origin=c]{-45}{MultiNews}
        & \rotatebox[origin=c]{-45}{TREC} & \rotatebox[origin=c]{-45}{TriviaQA} & \rotatebox[origin=c]{-45}{SAMSum}
        & \rotatebox[origin=c]{-45}{PRe}
        & \rotatebox[origin=c]{-45}{Lcc} & \rotatebox[origin=c]{-45}{RB-P}
        & \\
        
        \midrule
        \multicolumn{16}{c}{\large \textbf{Dream-v0-Instruct-7B}} \\
        \midrule
        \multicolumn{16}{c}{\small Full KV Cache} \\
        \midrule
        dLLM w/o Cache & 28.17 & 36.23 & 27.65 & 32.43 & 11.83 & 5.04 & 14.29 & 5.95 & 73.00 & 89.25 & 37.84 & 16.92 & 38.91 & 45.08 & 33.04 \\
        dLLM w/ Cache & 26.55 & 39.86 & 27.66 & 32.09 & 11.12 & 4.40 & 13.89 & 5.51 & 73.50 & 89.59 & 36.07 & 12.05 & 39.88 & 45.57 & 32.70 \\
        \midrule
        \multicolumn{16}{c}{\small B=16} \\
        \midrule
        SnapKV & \textbf{15.36} & 18.35 & 11.98 & 9.92 & 5.27 & 2.56 & 10.88 & 1.91 & 26.50 & 27.14 & 19.42 & 3.56 & 17.28 & 16.08 & 13.30 \\
        PyramidKV & 10.81 & 12.81 & 12.82 & 6.40 & 3.95 & \textbf{2.57} & 10.55 & 1.89 & 26.50 & 8.05 & 17.98 & 0.00 & 15.54 & 15.19 & 10.36 \\
        SqueezeAttention & 12.39 & 16.36 & 11.21 & 6.25 & 3.95 & 2.54 & 10.60 & 1.87 & 26.50 & 10.26 & 18.35 & 3.06 & 13.66 & 13.60 & 10.76 \\
        AdaKV & 15.31 & 15.14 & 12.18 & 8.50 & 6.44 & 2.54 & 10.79 & 1.82 & 26.50 & 21.56 & 20.08 & 5.11 & 16.91 & 17.10 & 12.86 \\
        \rowcolor{lightgreen} \textbf{MaskKV (Ours)} & 12.63 & \textbf{25.72} & \textbf{23.87} & \textbf{20.99} & \textbf{8.53} & 2.56 & \textbf{11.15} & \textbf{2.02} & \textbf{27.50} & \textbf{79.07} & \textbf{24.30} & \textbf{13.50} & \textbf{20.57} & \textbf{21.40} & \textbf{20.99} \\
        \midrule
        \multicolumn{16}{c}{\small B=32} \\
        \midrule
        SnapKV & 17.75 & 26.82 & 22.39 & \textbf{27.91} & 7.53 & \textbf{2.65} & \textbf{12.58} & 1.95 & 28.25 & 66.14 & 23.67 & 17.50 & 22.75 & 23.87 & 21.55 \\
        PyramidKV & 14.55 & 24.51 & 22.41 & 15.27 & 6.90 & 2.62 & 11.89 & 2.15 & 28.00 & 57.55 & 24.17 & 11.50 & 22.02 & 22.22 & 18.98 \\
        SqueezeAttention & 17.62 & 30.44 & 20.07 & 26.15 & 7.42 & 2.62 & 11.83 & 2.19 & 26.25 & 72.17 & 24.52 & 17.00 & 20.91 & 23.41 & 22.00 \\
        AdaKV & 17.09 & 25.42 & 23.64 & 24.77 & 7.55 & 2.58 & 12.46 & 1.89 & 27.75 & 69.77 & 25.21 & 18.83 & 21.67 & 24.98 & 21.69 \\
        \rowcolor{lightgreen} \textbf{MaskKV (Ours)} & \textbf{18.53} & \textbf{33.76} & \textbf{32.92} & 24.53 & \textbf{11.02} & 2.59 & 11.98 & \textbf{2.25} & \textbf{33.25} & \textbf{86.47} & \textbf{27.55} & \textbf{20.00} & \textbf{24.89} & \textbf{27.40} & \textbf{25.51} \\
        \midrule
        \multicolumn{16}{c}{\small B=64} \\
        \midrule
        SnapKV & 19.61 & 34.91 & 31.26 & 28.17 & 10.51 & 2.65 & 12.81 & 2.28 & 34.25 & 81.68 & 27.56 & \textbf{24.00} & 28.07 & 30.02 & 26.27 \\
        PyramidKV & 16.33 & 29.06 & 24.28 & 18.74 & 8.61 & \textbf{2.70} & 12.73 & 2.60 & 33.67 & 72.43 & 26.87 & 23.00 & 25.47 & 26.78 & 23.09 \\
        SqueezeAttention & 17.62 & 30.44 & 23.42 & 26.15 & 8.75 & 2.62 & 12.58 & 2.19 & 26.25 & 72.17 & 24.52 & 17.00 & 20.91 & 23.41 & 22.00 \\
        AdaKV & 20.98 & 36.31 & 33.10 & 29.06 & 9.55 & 2.59 & \textbf{13.12} & 2.29 & 35.25 & 85.24 & 28.50 & \textbf{24.00} & 28.09 & \textbf{32.76} & 27.20 \\
        \rowcolor{lightgreen} \textbf{MaskKV (Ours)} & \textbf{22.73} & \textbf{37.98} & \textbf{36.47} & \textbf{31.03} & \textbf{12.26} & 2.68 & 12.39 & \textbf{2.77} & \textbf{45.00} & \textbf{87.36} & \textbf{29.98} & 22.50 & \textbf{29.11} & 31.61 & \textbf{28.85} \\
        \midrule
        \multicolumn{16}{c}{\small B=128} \\
        \midrule
        SnapKV & \textbf{22.36} & 39.02 & 30.28 & 32.27 & 11.51 & 2.72 & \textbf{13.20} & 2.72 & 38.00 & 87.17 & 28.69 & \textbf{26.05} & \textbf{32.55} & 36.51 & 28.79 \\
        PyramidKV & 17.31 & 36.59 & 25.21 & 24.38 & 10.99 & 2.77 & 12.75 & 2.90 & 39.75 & 84.83 & 29.71 & 24.00 & 30.26 & 30.33 & 26.56 \\
        SqueezeAttention & 19.36 & 36.19 & 28.35 & 29.79 & 10.25 & 2.81 & 13.11 & 2.64 & 33.00 & 85.99 & 29.25 & 25.00 & 27.11 & 31.92 & 26.77 \\
        AdaKV & 21.97 & \textbf{39.68} & 33.24 & 31.57 & 11.38 & 2.69 & 13.15 & 2.80 & 38.75 & 88.51 & 30.43 & 24.25 & 31.46 & \textbf{37.28} & 29.08 \\
        \rowcolor{lightgreen} \textbf{MaskKV (Ours)} & 21.49 & 39.23 & \textbf{38.25} & \textbf{33.99} & \textbf{14.76} & \textbf{2.86} & 12.48 & \textbf{3.19} & \textbf{53.50} & \textbf{88.69} & \textbf{30.97} & 21.92 & 30.71 & 34.79 & \textbf{30.49} \\
        \midrule
        \multicolumn{16}{c}{\small B=256} \\
        \midrule
        SnapKV & 23.01 & 40.81 & 35.52 & 33.92 & 12.08 & 2.89 & 13.06 & 3.33 & 44.75 & 89.57 & 32.16 & 21.65 & \textbf{35.49} & 37.61 & 30.42 \\
        PyramidKV & 21.50 & 36.74 & 28.35 & 29.08 & 9.17 & 2.79 & 13.06 & 3.55 & 46.17 & 88.44 & 31.74 & 22.25 & 32.78 & 33.09 & 28.55 \\
        SqueezeAttention & 21.82 & 41.20 & 35.80 & 33.23 & 11.36 & 2.83 & 13.11 & 2.56 & 37.00 & \textbf{89.63} & 30.79 & \textbf{25.00} & 30.29 & 34.69 & 29.24 \\
        AdaKV & 24.00 & \textbf{43.32} & 35.86 & 33.75 & 11.97 & 2.90 & \textbf{13.14} & 3.41 & 48.50 & 87.70 & 32.66 & 19.96 & 35.40 & \textbf{39.41} & 30.86 \\
        \rowcolor{lightgreen} \textbf{MaskKV (Ours)} & \textbf{24.13} & 39.43 & \textbf{36.47} & \textbf{31.93} & \textbf{12.66} & \textbf{2.93} & 12.90 & \textbf{3.74} & \textbf{58.00} & 88.52 & \textbf{32.93} & 20.58 & 32.22 & 35.94 & \textbf{30.88} \\
        \midrule
        \multicolumn{16}{c}{\small B=512} \\
        \midrule
        SnapKV & 24.27 & 40.23 & 35.32 & 34.87 & 11.65 & 2.91 & 13.32 & 4.15 & 51.00 & \textbf{89.66} & 33.09 & 16.29 & 37.22 & 40.47 & 31.03 \\
        PyramidKV & 21.50 & 36.74 & 28.35 & 29.08 & 9.17 & 2.79 & 13.06 & 3.55 & 46.17 & 88.44 & 31.74 & \textbf{22.25} & 32.78 & 33.09 & 28.55 \\
        SqueezeAttention & 22.17 & 40.15 & 35.98 & \textbf{35.63} & 11.32 & 2.89 & 13.19 & 3.91 & 45.00 & 86.95 & 32.37 & 17.08 & 33.04 & 38.75 & 29.89 \\
        AdaKV & 23.86 & \textbf{40.62} & \textbf{37.45} & 34.66 & 12.58 & 3.05 & \textbf{13.61} & 3.81 & 57.75 & 89.15 & 33.50 & 17.42 & \textbf{37.23} & \textbf{42.35} & \textbf{31.93} \\
        \rowcolor{lightgreen} \textbf{MaskKV (Ours)} & \textbf{25.90} & 39.71 & 33.27 & 34.01 & \textbf{13.68} & \textbf{3.10} & 12.92 & \textbf{4.26} & \textbf{60.75} & 87.49 & \textbf{33.78} & 16.81 & 34.19 & 38.83 & 31.34 \\

        \bottomrule
    \end{tabular}
    } 
\end{table*}

\begin{table}[t!]
\caption{Comparison of \textbf{dLLM-Cache} with its simplified variants.}
\label{tab:ablation_memory}
\centering
\begin{tabular}{l c}
\toprule
\textbf{Method} & \textbf{Memory (GB)} \\
\midrule
\textbf{dLLM-Cache}                            & 68.13 \\
\midrule
\quad + Mask-Only Projection                 & 53.74 \\
\quad + Prompt state exclusion             & 38.12 \\
\quad + \mymethod (online)                            & 38.12 \\
\quad + \mymethod (offline)                            & \textbf{23.42} \\
\bottomrule
\end{tabular}
\end{table}

\subsection{Details of \emph{MaskKV}}
\label{sec:maskkv_details}
\paragraph{Prompt-state Exclusion.}
In dLLM-Cache, features from both the prompt and response tokens (including keys, values, attention outputs, and MLP activations) are cached at each denoising step.
However, we observe that only the key–value representations of prompt tokens contribute to the attention computation of mask tokens, while the prompt-side attention and MLP outputs have no downstream influence.
We therefore exclude these redundant prompt features from caching and retain only their key–value pairs, which substantially reduces memory usage without affecting generation quality.

\paragraph{Mask-only Projection.}
In the official LLaDA implementation, after the final layer computation, the model projects all tokens (including both prompt and mask positions) into the vocabulary space to produce logits.
This operation yields a large but unnecessary tensor, as the logits of prompt tokens are never used during decoding.
We thus restrict the vocabulary projection to masked positions only and skip the prompt ones.
This \emph{mask-only projection} optimization removes redundant matrix multiplications and further reduces GPU memory consumption.

\subsection{Implementation Details}
\label{implement details}
\paragraph{Evaluation Metrics.}
We evaluate both the efficiency and quality of our method using quantitative metrics. 
Generation quality is assessed with the official task-specific metrics (see Table~\ref{tab:longbench_datasets_simplified} for details) of LongBench, which measure model accuracy under cache eviction.  
Computation efficiency is reported in Tokens Per Second (TPS), reflecting the average number of tokens decoded per second.  
For memory efficiency, we track both the peak GPU memory during inference and the size of the KV cache.  
Under bf16 precision, the KV cache memory footprint is given by  
\begin{equation}
\text{Mem}_{\text{KV}} = 2 \cdot L \cdot H \cdot d_{\text{head}} \cdot s_{\text{bf16}},
\end{equation}
where $L$ is the sequence length, $H$ the number of attention heads, $d_{\text{head}}$ the dimension per head, and $s_{\text{bf16}}=2$ bytes denotes the storage size of a \texttt{bf16} element.  
The factor $2$ represents the storage requirements for both key and value states.

\paragraph{Baseline.}
We compare one token-selection scheme and three architectural budget-allocation policies under an identical cache budget, enabling a fair, apples-to-apples assessment of their effectiveness. \\
For the token selection strategy, we evaluate one strong KV-cache compression method as a baseline for autoregressive models.
\begin{itemize}
  \item \textbf{SnapKV} uses a small ``observation window'' at the end of the prompt to predict which parts of the entire context are most important. It analyzes the attention scores from this window in ``voting'' mechanism to identify and select these key parts.
\end{itemize}
For architectural budget allocation, we evaluate competitive approaches that distribute the budget across the model's various structural components.
\begin{itemize}
  \item \textbf{PyramidKV} implements a static, non-uniform budget allocation where the cache capacity of each layer is a direct function of its depth. This function is engineered to be monotonically decreasing, granting the maximal budget to the lowest layers and progressively constricting it for higher layers that process more semantically aggregated representations.
  \item \textbf{SqueezeAttention} gauges layer importance by calculating the cosine similarity between the input and output of an attention block. Based on this score, it classifies layers into three tiers and assigns a minimal cache budget to the least important one.
  \item \textbf{Ada-KV} allocates its cache budget in a fine-grained, adaptive manner: it first assesses the relative importance of each key–value (KV) pair across all attention heads, then distributes the budget proportionally, granting a larger share of resources to KVs belonging to the most salient heads.
\end{itemize}

\paragraph{Parameters.}
Our experimental parameters are configured in accordance with prior research\citep{li2024snapkv, cai2024pyramidkv,wang2024squeezeattention, feng2024ada}. The specific settings are as follows:

\begin{itemize}
    \item \textbf{Default Selection Method}: Unless otherwise specified, we adopt \textbf{SnapKV} as the foundational selection method for all budget allocation strategies. For SnapKV~\citep{li2024snapkv} itself, the final window size is set to \textbf{32}, consistent with its application in the LongBench benchmark.
    \item \textbf{Pyramid-based Allocation}: We set the hyperparameter $\beta$, which directly controls the ``steepness''
    of the allocation pyramid, to \textbf{20}, adhering to the default value proposed in the original paper~\citep{cai2024pyramidkv}.
    \item \textbf{SqueezeAttention}: We cluster the layers into three distinct groups. A \textbf{40\%} budget is allocated to the least important group, a setting identified as optimal in its original study.
    \item \textbf{AdaKV}\citep{feng2024ada}: We reserve a \textbf{20\%} budget for uniform allocation. This measure is implemented to prevent the assignment of excessively small budgets to highly sparse attention heads.
\end{itemize}

\paragraph{Experiment settings.}
To ensure reproducibility, we outline our experimental settings. Unless otherwise specified, our default configuration sets the prompt refresh interval to 50, the response refresh interval to 5, the transfer ratio to 0.25, and the block length to 8. The step size is set equal to the generation length, which is specified for each task in Table~\ref{tab:longbench_datasets_simplified}.
For the results presented in Fig.~\ref{fig:across_size} and the ablation study in Tab.~\ref{tab:ablation}, the KV cache budget is set to 256. The experiment in Fig.~\ref{fig:hy}~ is conducted on the HotpotQA dataset with a budget of 32. For the NIAH baseline, we adopt the same configuration as that used in DuoAttention~\citep{xiao2024duoattention}.

\paragraph{Datasets.}

We conducted evaluations using LongBench~\citep{bai2023longbench}.
The LongBench benchmark \citep{bai2023longbench} evaluates large language models across a diverse set of long-context tasks. The benchmark is structured into six key domains:
\begin{itemize}
    \item \textbf{Single-Document QA:} Assesses a model's ability to extract answers from a single source document. This category utilizes datasets such as NarrativeQA~\citep{kovcisky2018narrativeqa}, Qasper~\citep{dasigi2021dataset}, and MultiFieldQA~\citep{bai2023longbench}, covering documents ranging from academic papers and legal files to encyclopedias.

    \item \textbf{Multi-Document QA:} Challenges models to synthesize information from multiple documents to formulate a coherent answer. It employs Wikipedia-based multi-hop QA datasets, including HotpotQA~\citep{yang2018hotpotqa}, 2WikiMultihopQA~\citep{ho2020constructing}, and MuSiQue~\citep{trivedi2022musique}.

    \item \textbf{Summarization:} Tests a model's capacity for comprehensive understanding and condensation of long texts. The datasets for this task are GovReport~\citep{huang2021efficient}, QMSum~\citep{zhong2021qmsum}, and the multi-document corpus MultiNews~\citep{fabbri2019multi}.

    \item \textbf{Few-shot Learning:} Measures a model's adaptability on a variety of tasks with limited examples. This includes classification with TREC~\citep{li2002learning}, conversational summarization with SAMSum~\citep{gliwa2019samsum}, and reading comprehension with TriviaQA~\citep{joshi2017triviaqa}.
    
    \item \textbf{Synthetic Tasks:} Purpose-built challenges designed to test specific abilities, such as counting unique passages (PassageCount~\citep{bai2023longbench}) or matching a summary to its source passage (PassageRetrieval-en~\citep{raffel2020exploring}).

    \item \textbf{Code Completion:} Evaluates a model's proficiency in generating code based on existing context. This is tested using the LCC~\citep{guo2023longcoder} dataset for single-file contexts and the RepoBench-P~\citep{liu2023repobench} dataset for tasks requiring information aggregation across multiple files.
\end{itemize}
\begin{table*}
    \centering
    \caption{Detailed information of the datasets in the LongBench benchmark.}
    \label{tab:longbench_datasets_simplified}
    \setlength{\tabcolsep}{5pt} 
    \begin{tabular}{lllrcrcr}
        \toprule
        \textbf{Label} & \textbf{Task} & \textbf{Eval Metric} & \textbf{Avg Len} & \textbf{Gen Len} & \textbf{Language} & \textbf{Sample Num} \\
        \midrule
        NrtvQA & NarrativeQA & F1 & 18,409 & 128 & EN & 200 \\
        Qasper & Qasper & F1 & 3,619 & 128 & EN & 200 \\
        MF-en & MultiFieldQA-en & F1 & 4,559 & 64 & EN & 150 \\
        HotpotQA & HotpotQA & F1 & 9,151 & 32 & EN & 200 \\
        2WikiMQA & 2WikiMultihopQA & F1 & 4,887 & 32 & EN & 200 \\
        Musique & MuSiQue & F1 & 11,214 & 32 & EN & 200 \\
        GovReport & GovReport & Rouge-L & 8,734 & 512 & EN & 200 \\
        QMSum & QMSum & Rouge-L & 10,614 & 512 & EN & 200 \\
        MultiNews & MultiNews & Rouge-L & 2,113 & 512 & EN & 200 \\
        TREC & TREC & Accuracy & 5,177 & 64 & EN & 200 \\
        TriviaQA & TriviaQA & F1 & 8,209 & 32 & EN & 200 \\
        SAMSum & SAMSum & Rouge-L & 6,258 & 128 & EN & 200 \\
        PCount & PassageCount & Accuracy & 11,141 & 32 & EN & 200 \\
        PRe & PassageRetrieval-en & Accuracy & 9,289 & 32 & EN & 200 \\
        Lcc & LCC & Edit Sim & 1,235 & 64 & Python/C\#/Java & 500 \\
        RB-P & RepoBench-P & Edit Sim & 4,206 & 64 & Python/Java & 500 \\
        \bottomrule
    \end{tabular}
\end{table*}


\subsection{A Formal Proof on the Primacy of Mask Attention}
\label{the primacy of mask attention}
\subsubsection{Preliminaries and Notation}

To ensure the rigor of the proof, we first define the symbols and notation used throughout this section.
\begin{itemize}
    \item \textbf{Input Sequence:} The input sequence $X \in \mathbb{R}^{n \times d}$ consists of embeddings for $n$ tokens, where $d$ is the embedding dimension. The sequence $X$ is partitioned into two parts:
    \begin{itemize}
        \item \textbf{Prompt:} $X_p \in \mathbb{R}^{n_p \times d}$, with its set of token indices denoted as $S_p$.
        \item \textbf{Mask:} $X_m \in \mathbb{R}^{n_m \times d}$, with its set of token indices denoted as $S_m$.
    \end{itemize}
    The full sequence is a concatenation $X = [X_p, X_m]$, with a total length of $n = n_p + n_m$.

    \item \textbf{Transformer Layer:} A standard Transformer model consists of $L$ identical layers stacked on top of each other. Let $h^{(l)} \in \mathbb{R}^{n \times d}$ denote the output hidden state of the $l$-th layer, where $l \in \{1, \dots, L\}$. We define the initial embedding as the output of the 0-th layer, i.e., $h^{(0)} = X$.

    \item \textbf{Intra-Layer Computation:} The computation within layer $l$ can be represented as a function, $\text{Block}$, which takes the output of the previous layer $h^{(l-1)}$ as input:
    $$h^{(l)} = \text{Block}(h^{(l-1)})$$
    To analyze the information flow, we can abstract the update process of each layer. The output of layer $l$ is the sum of its input and an update term $\Delta^{(l)}$:
    $$h^{(l)} = h^{(l-1)} + \Delta^{(l)}$$
    where $\Delta^{(l)}$ represents the total update contributed by the sub-layers (MHA and FFN) of layer $l$.
\end{itemize}

\begin{table}[t!]
\caption{Comparison of token selection strategies on \textbf{gsm8k} with \textbf{budget=128}.}
\label{tab:voting}
\centering
\begin{tabular}{l c}
\toprule
\textbf{Token Selection Strategy} & \textbf{Score (\%)} \\
\midrule
SnapKV       & 64.90 \\
Prompt-Voting  & 60.35 \\
Mask-Voting    & \textbf{68.08} \\
All-Voting     & 66.64 \\
\bottomrule
\end{tabular}
\end{table}

\subsubsection{Proposition}

For any mask token $m \in S_m$, its final hidden state $h^{(L)}_m$, which directly determines the predictive logits, can be precisely expressed as the sum of its initial embedding $h^{(0)}_m$ and the cumulative updates from all $L$ layers of the model. Within these updates, the Multi-Head Attention (MHA) mechanism serves as the sole channel for the mask token to incorporate information from the prompt tokens. Consequently, the attention scores originating from mask queries are the most direct and fundamental indicators of the importance of prompt information for the model's generative process.

\begin{table*}[t!]
    \centering
    \small
    \caption{Effect of mask token position on voting performance (\textbf{budget=256}). The best score in each column and the best overall average are highlighted in \textbf{bold}.}
    \label{tab:mask_voting_position}
    \setlength{\tabcolsep}{5pt}
    \renewcommand{\arraystretch}{1.1}

    \begin{tabular}{@{}lcccccc@{}}
        \toprule
        \textbf{Mask Position} &
        \textbf{HotpotQA} &
        \textbf{2WikiMQA} &
        \textbf{Musique} &
        \textbf{TriviaQA} &
        \textbf{PRe} &
        \textbf{Average} \\
        \midrule
        first (front) & 12.39 & 15.72 & 9.19 & 41.69 & \textbf{99.75} & 35.75 \\
        second (middle) & 12.89 & 15.85 & 8.49 & 37.99 & 96.58 & 34.36 \\
        third (middle) & 12.82 & \textbf{16.82} & 8.72 & 39.74 & 96.00 & 34.82 \\
        last (back) & \textbf{14.33} & 15.97 & \textbf{9.69} & \textbf{44.75} & 97.50 & \textbf{36.45} \\
        \bottomrule
    \end{tabular}
\end{table*}

\subsubsection{Proof}

The proof proceeds in three steps. First, we establish the central role of $h_m^{(L)}$ by considering the model's objective function. Second, we derive the compositional structure of $h_m^{(L)}$ through a recursive expansion. Finally, we analyze the components of this structure to demonstrate the unique role of the attention mechanism.

\paragraph{Step 1: The Inference Objective and Decisive Computations}
During inference, the objective of the model is to predict a sequence of tokens for the positions specified by the mask index set, $S_m$. This generative process begins with the computation of the final hidden states, $h^{(L)} \in \mathbb{R}^{n \times d}$, for the entire sequence. The language model head (LM Head), a linear projection matrix $W_{out} \in \mathbb{R}^{d \times |V|}$ (where $|V|$ is the vocabulary size), then maps these hidden states to logit vectors:
$$ \text{Logits} = h^{(L)} \cdot W_{out} $$
A ``softmax'' function is subsequently applied to the logits at each position to yield a probability distribution over the vocabulary.

Critically, for the task at hand, our interest lies exclusively in the logits at the active mask positions ($S_m$), since all other tokens—whether part of the original prompt ($S_p$) or already unmasked in prior steps—are considered fixed context. Therefore, the generative process, whether it be greedy decoding or sampling, is performed exclusively on the probability distributions corresponding to the mask positions. This implies that the quantities of interest, which solely determine the generated output, are the final hidden states of the mask tokens, $\{h_m^{(L)} \mid m \in S_m\}$. 

\paragraph{Step 2: Recursive Expansion of the Hidden State}
Based on the abstract update rule $h^{(l)} = h^{(l-1)} + \Delta^{(l)}$, we can perform a recursive expansion (a telescoping sum) for the final hidden state $h_m^{(L)}$ of any mask token $m$:
\begin{align*}
    h_m^{(L)} &= h_m^{(L-1)} + \Delta_m^{(L)} \\
              &= (h_m^{(L-2)} + \Delta_m^{(L-1)}) + \Delta_m^{(L)} \\
              &= h_m^{(L-2)} + \Delta_m^{(L-1)} + \Delta_m^{(L)} \\
              &\vdots \\
              &= h_m^{(0)} + \sum_{l=1}^{L} \Delta_m^{(l)} \label{eq:telescoping_sum}
\end{align*}
This expansion is the mathematical centerpiece of our proof, showing that the final representation is an accumulation of updates upon its initial state.

\paragraph{Step 3: Analysis of the Update Components}
We now analyze the composition of the cumulative update term $\sum_{l=1}^{L} \Delta_m^{(l)}$. Each layer's update $\Delta_m^{(l)}$ consists of contributions from the MHA and FFN sub-layers: $\Delta_m^{(l)} = \text{MHA}_m^{(l)} + \text{FFN}_m^{(l)}$.
\begin{itemize}
    \item \textbf{Contribution of the Feed-Forward Network (FFN):} The FFN is a position-wise transformation. Its computation for token $m$ is independent of all other tokens $j \neq m$. Thus, the FFN can only process and non-linearly transform the information already present in $h_m$; it cannot introduce new information from the prompt.

    \item \textbf{Contribution of Multi-Head Attention (MHA):} The MHA mechanism is fundamentally different. The output for token $m$, $\text{AttnOut}_m$, is a weighted sum of the Value vectors $V_j$ of all tokens in the sequence:
    $$ \text{AttnOut}_m = \sum_{j=1}^{n} \alpha_{mj} V_j $$
    where the attention weight $\alpha_{mj} = \text{softmax}\left(\frac{Q_m K_j^T}{\sqrt{d_k}}\right)$. The summation index $j$ spans all tokens, including those in the prompt ($j \in S_p$). This demonstrates that MHA is the \textbf{exclusive} mechanism that allows for information exchange between different token positions. For a mask token $m \in S_m$, only through MHA can it interact with prompt tokens $j \in S_p$ to aggregate relevant information.
\end{itemize}
Combining these points, we see that within the final representation $h_m^{(L)} = h_m^{(0)} + \sum_{l=1}^{L} (\text{MHA}_m^{(l)} + \text{FFN}_m^{(l)})$, the MHA term is the sole channel through which information from the prompt can be incorporated into the mask token's representation.

\subsubsection{Conclusion and Implication}

In conclusion, our theoretical proof establishes the primacy of attention guided by mask queries within the generative paradigm of dLLMs. By demonstrating that this mechanism is the indispensable information bridge from context to target, our findings provide a robust theoretical foundation for novel inference strategies, such as the KV cache selection method proposed in this work.

\end{document}